\pdfoutput=1

\documentclass[10pt,letterpaper]{article}

\usepackage[top=0.85in,left=2.75in,footskip=0.75in]{geometry}
\usepackage{amsmath,amssymb}        % amsmath and amssymb packages, useful for mathematical formulas and symbols
\usepackage{changepage}             % Use adjustwidth environment to exceed column width (see example table in text)
\usepackage[utf8x]{inputenc}        % Use Unicode characters when possible
\usepackage{textcomp,marvosym}      % textcomp package and marvosym package for additional characters
\usepackage{cite}                   % cite package, to clean up citations in the main text. Do not remove.
\usepackage{nameref,hyperref}       % Use nameref to cite supporting information files (see Supporting Information section for more info)
\usepackage{microtype}              % ligatures disabled
\DisableLigatures[f]{encoding = *, family = * }

\usepackage[table]{xcolor}          % color can be used to apply background shading to table cells only
\usepackage{array}                  % array package and thick rules for tables

\newcolumntype{+}{!{\vrule width 2pt}}  

\newlength\savedwidth

\raggedright
\usepackage[aboveskip=1pt,labelfont=bf,labelsep=period,justification=raggedright,singlelinecheck=off]{caption}

\makeatletter
\renewcommand{\@biblabel}[1]{\quad#1.}
\makeatother

\usepackage{lastpage,fancyhdr,graphicx}
\usepackage{epstopdf}
\fancyheadoffset[L]{2.25in}
\fancyfootoffset[L]{2.25in}
\usepackage{ragged2e}
\usepackage{appendix}
\usepackage{cjhebrew}
\usepackage{color,soul}

\begin{document}
\vspace*{0.2in}

\begin{flushleft}
{\Large
\textbf\newline{Artificial intelligence based writer identification generates new evidence for the unknown scribes of the Dead Sea Scrolls exemplified by the Great Isaiah Scroll (1QIsa\textsuperscript{a})} 
}
\newline
\\
Mladen Popovi\'{c}\textsuperscript{1\Yinyang *},
Maruf A. Dhali \textsuperscript{2\Yinyang},
Lambert Schomaker\textsuperscript{2\Yinyang}
\\
\bigskip
\textbf{1} Qumran Institute, Faculty of Theology and Religious Studies, University of Groningen, Groningen, The Netherlands
\\
\textbf{2} Department of Artificial Intelligence, Faculty of Science and Engineering, University of Groningen, Groningen, The Netherlands
\\
\bigskip

\Yinyang \ These authors contributed equally to this work.\\
\bigskip

* Corresponding author's address: Qumran Institute, Oude Boteringestraat 38, 9712 GK Groningen, The Netherlands. Email: m.popovic@rug.nl
\end{flushleft}

\justify

\section*{Abstract}
The Dead Sea Scrolls are tangible evidence of the Bible’s ancient scribal culture. Palaeography—the study of ancient handwriting—can provide access to this scribal culture. However, one of the problems of traditional palaeography is to determine writer identity when the writing style is near uniform. This is exemplified by the Great Isaiah Scroll (1QIsa\textsuperscript{a}). To this end, we used pattern recognition and artificial intelligence techniques to innovate the palaeography of the scrolls regarding writer identification and to pioneer the microlevel of individual scribes to open access to the Bible’s ancient scribal culture. Although many scholars believe that 1QIsa\textsuperscript{a} was written by one scribe, we report new evidence for a breaking point in the series of columns in this scroll. Without prior assumption of writer identity, based on point clouds of the reduced- dimensionality feature-space, we found that columns from the first and second halves of the manuscript ended up in two distinct zones of such scatter plots, notably for a range of digital palaeography tools, each addressing very different featural aspects of the script samples. In a secondary, independent, analysis, now assuming writer difference and using yet another independent feature method and several different types of statistical testing, a switching point was found in the column series. A clear phase transition is apparent around column 27. Given the statistically significant differences between the two halves, a tertiary, post-hoc analysis was performed by visual inspection of character heatmaps and of the most discriminative fraglet sets in the script. Demonstrating that two main scribes were responsible for the Great Isaiah Scroll, this study sheds new light on the Bible’s ancient scribal culture by providing new, tangible evidence that ancient biblical texts were not copied by a single scribe only but that multiple scribes could closely collaborate on one particular manuscript.

% \linenumbers
\section{Introduction}\label{section:Introductio}
Ever since their modern discovery, the Dead Sea Scrolls are famous for containing the oldest manuscripts of the Hebrew Bible (Old Testament) and many hitherto unknown ancient Jewish texts. The manuscripts date from the 4\textsuperscript{th} century BCE to the 2\textsuperscript{nd} century CE. They come from the caves near Qumran and other Judaean Desert sites west near the Dead Sea, except for Wadi Daliyeh which is north of Jericho \cite{popovic2019}. Among other things, the scrolls provide a unique vantage point for studying the latest literary evolutionary phases of what were to become the Hebrew Bible. As archaeological artifacts, they offer tangible evidence for the Bible's ancient scribal culture `in action'.

A crucial but yet hardly used entry point into the Bible’s ancient scribal culture is that of individual scribes \cite{zahn2017beyond}. There is, however, a twofold problem with putting this entry point of individual scribes into effective use. Except for a handful of named scribes in a few documentary texts \cite{cotton1997discoveries,wise2015language}, the scribes behind the scrolls are anonymous. This is especially true for the scrolls from Qumran, which, with almost a thousand manuscripts of mostly literary texts, represents the largest find site. 

The next best thing to scribes identified by name is scribes identified by their handwriting. Although some of the suggestions for a change of scribal hands in a single manuscript or scribes who copied more than one manuscript \cite{tov2004scribal,tigchelaar2003search,yardeni2007note,ulrich2007identification} have met with broader assent, most have not been assessed at all. And estimations of the total number of scribes \cite{allegro1957dead,wise1994accidents,alexander2003literacy}, an argument in the discussion about the origin of the scrolls from Qumran \cite{wise2015language,golb1990khirbet,golb1995wrote,crawford2019scribes}, have been, at best, educated guesses.

One of the main problems regarding traditional palaeography of the Dead Sea Scrolls, and also for writer identification in general \cite{sirat2007writing,papaodysseus2014identifying}, is the ability to distinguish between variability within the writing of one writer and similarity in style—but with subtle variations—between different writers. On the one hand, scribes may show a range in a variety of forms of individual letters in one or more manuscripts. On the other hand, different scribes may write in almost precisely the same way, making it a challenge to determine the individual scribe beyond similarities in the general style. 

The question is whether perceived differences in handwriting are significant and the result of there being two different writers or insignificant because they are the result of normal variations within the handwriting of the same writer. The problem with knowing which differences are likely to be idiographic, and thus significant, is that, in the end, this also involves using implicit criteria that are experience-based \cite{sirat2007writing,davis2007practice}. In this regard, although they work according to differing methodologies \cite{sirat2007writing,davis2007practice,harralson2017huber}, there is no difference between professional forensic document examiners and palaeographers. The problem is also how one can convince others \cite{derolez2003palaeography,stokes2015digital}, whether through pictorial form, verbal descriptions, palaeographic charts or a combination thereof. 

The Great Isaiah Scroll from Qumran Cave 1 (1QIsa\textsuperscript{a}) exemplifies the lack of a robust method in Dead Sea Scrolls palaeography for how to determine and verify writer identity or difference, especially when the handwriting is near uniform. The question for 1QIsa\textsuperscript{a} is whether subtle differences in writing should be regarded as normal variations in the handwriting of one scribe or as similar scripts of two different scribes and, if the latter, whether the writing of the two scribes coincides with the two halves of the manuscript. The scroll measures 7.34 m in length, averages 26 cm in height, and contains 54 columns of Hebrew text. There is a codicological caesura between columns 27 and 28 in the form of a three-line lacuna at the bottom of column 27. In the second half of the scroll the orthography and morphology of the Hebrew is different and there are spaces left blank.  

Scholars have perceived an almost uniform writing style throughout the manuscript of 1QIsa\textsuperscript{a} \cite{kahle1950present,kahle1951hebraischen}, yet also acknowledged that different scribes could have shared a similar writing style \cite{noth1951note,kuhl1952clerkproperty}—the script type is called Hasmonaean in the field, the style of writing is formal, and the manuscript is traditionally dated to the late 2\textsuperscript{nd} century BCE. Accordingly, some have argued that two scribes were each responsible for copying half of the manuscript, columns 1–27 and columns 28–54 \cite{tov2004scribal,brookebisection}. But most scholars have argued or assumed that the entire manuscript was copied by one scribe, with minor interventions by other, contemporaneous and also much later, scribes \cite{trever1949paleographic,ulrich2011discoveries,justnes2015hand}, and that differences between the two halves should be explained otherwise, for example, by assuming that two separate and dissimilar \textit{Vorlagen} were used or that the \textit{Vorlage} for the second half was a damaged manuscript \cite{brownlee1952manuscripts,martin1958scribal,kutscher1974language,giese1988further,cook1989orthographical,cook1992dichotomy,pulikottil2001transmission,williamsonscribe,longacre2013developmental}. 

No one, however, has provided detailed palaeographic arguments for writer identity or difference in 1QIsa\textsuperscript{a}, except for \cite{ulrich2011discoveries} who provided a palaeographic chart to argue for one main scribe. But the palaeographic chart in \cite{ulrich2011discoveries} is insufficient to demonstrate this for at least three reasons (additional details about the supposed scribal idiosyncrasies are provided in the supplementary material \ref{sec:appen1:scribal}). Having been electronically produced it is unclear where, and how exactly, the characters were taken from. It is unclear whether “the typical form of the letters” is deemed typical because it is the most common form or because it is idiographic, understood as a subtle variation in graphic form that gives evidence of individuality \cite{davis2007practice}. Finally, the crucial question is how large amounts of data were processed to generate the chart. The number of instances of a specific Hebrew letter may run in the thousands in 1QIsa\textsuperscript{a}. 

Here, pattern recognition and artificial intelligence techniques can assist researchers by processing large amounts of data and by producing quantitative analyses that are impossible for a human to perform. Over the years, within the field of pattern recognition, dedicated feature extraction techniques have been proposed and studied in identifying writers. By extracting useful quantitative data that is writer specific, these techniques are used on handwritten documents to produce feature vectors. In one of our earlier studies, we have tested both textural-based and grapheme-based features on a limited number of scrolls to identify scribes \cite{dhali2017digital}. Textural-based features use the statistical information of slant and curvature of the handwritten characters. Grapheme-based features extract local structures of characters and then map them into a common space, similar to the so-called bag-of-words approach in text analysis \cite{he2017beyond}. 

We have already shown that extracting Hinge, a textural feature operating on the microlevel of handwriting, can be useful in identifying writers \cite{bulacu2007text}. In the process of producing character shapes, writers subconsciously slow down and speed up their hand movements. For example, a bend within a character is an indication of where a slowing down took place, and the sharper the bend the greater the deceleration of the hand movement. Hinge uses this intuitive information between the static space and dynamic time to produce a feature vector.

Similar to the textural features, allographs (prototypical character shapes) can also be useful for writer identification \cite{niels2007automatic}. Allographs can be obtained from either the full characters or from part/s of the characters. We have already worked with full characters and used them to create a codebook of the Dead Sea Scrolls characters for style development analysis \cite{dhali2020feature}.

The quantitative evidence is additional evidence that can stimulate palaeographers to explicate their qualitative analyses \cite{stokes2015digital,ciula2005digital}. Pattern recognition and artificial intelligence techniques do not give certainty of identification but they give statistical probabilities that can help the human expert understand and also decide between the likelihood of different possibilities. 

\begin{figure}[!ht]
    \centering
    \includegraphics[width=.6\textwidth]{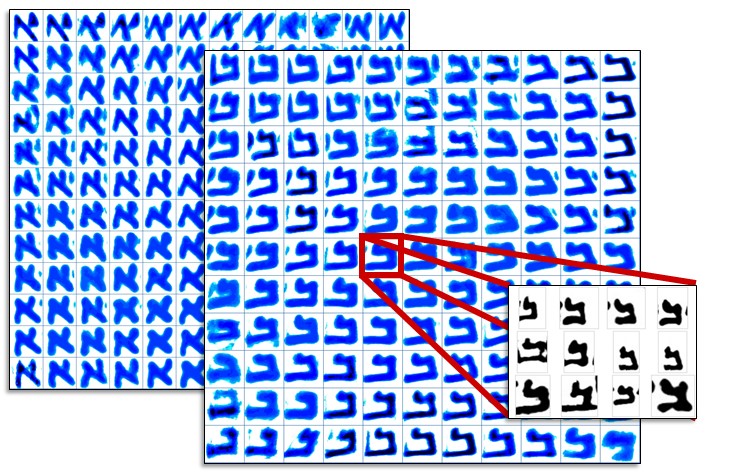}
	\caption{Two 12x12 Kohonen maps (blue colormaps) of (full) character \textit{aleph} and \textit{bet} from the DSS collection. Each of the characters in the Kohonen maps are formed from multiple instances of similar characters (shown with zoomed box with red lines). These maps are useful for chronological style development analysis. In our current study of writer identification, fraglets will be used instead of full character shapes to achieve more precise (robust) results.}
	\label{fig:intro1}
\end{figure}

The evidence from pattern recognition methods can be presented in numbers (quantification of distance; the choice of distance measures plays an important role) but also, more helpfully, in two- or three-dimensional visualizations. Also, so-called Kohonen self-organizing feature maps (see Fig.\ref{fig:intro1}) and heatmaps may prove important for detecting a typical style of a letter (a centroid) that is the computed average of all particular instances that were most similar to it. Although such a centroid statistically is a reliable attractor for shapes that look like it, its visual pattern may not consist of a particular canonical or idealized form. Inspection of the individual instances belonging to a centroid (i.e., its members) will reveal the characteristics of that cluster of shapes. Such analyses may supplement exhaustive letter-by-letter analysis.

Our research demonstrates that two main scribes can be identified in 1QIsa\textsuperscript{a} and that they coincide with columns 1–27 and columns 28–54. This study illustrates the advantage of the innovative use of robust pattern recognition and artificial intelligence techniques for writer identification in the Dead Sea Scrolls when dealing with an almost uniform writing style that makes it difficult, if not near impossible, for researchers to assess writer identity or difference. Moreover, we show that procedures for cross-examination \cite{davis2007practice,stokes2015digital} and falsification are in place by statistical and post-hoc visual analyses. Bridging artificial intelligence and traditional palaeography, our post-hoc visual analyses go beyond the state of the art by correlating the quantitative analyses to a level suitable for researchers to be able to see what the computer `sees’, enabling a new way of looking at palaeographic evidence. Also, our analysis is \textit{fully automatic}. We have no need to apply a semi-automatic first step of character reconstruction as in \cite{faigenbaum2016algorithmic,faigenbaum2020algorithmic,shaus2020forensic} that aim to imitate the ancient reed pen’s movement, although it seems more likely that the stiff-flexible fibrous tip of the sea-rush stem must have been used, like in Egypt \cite{KooijG2015}. We have developed robust and sufficiently delicate binarization and extraction methods and have succeeded at extracting the ancient ink traces as they appear on digital images \cite{dhali2019binet}. This is important because the ancient ink traces relate directly to a person’s muscle movement and are person specific. For writer identification one should ideally work with the original written content only. The pattern recognition and artificial intelligence techniques should therefore be capable of focusing on the original written content only. Neither should it depend on modern character reconstructions.

In a way that was not possible before, our approach opens access to the tangible evidence of the hitherto almost completely denied microlevel of the individual scribes of the Dead Sea Scrolls and the possibility to examine the different compositions copied by each of the scribes. The change of scribal hands in a literary manuscript or the identification of one and the same scribe in multiple manuscripts can be used as evidence to understand various forms of scribal collaboration that otherwise remain unknown to us. The number of literary manuscripts on which a scribe worked, either alone or with others, can serve as tangible evidence for understanding processes of textual and literary creation, circulation, and consumption. Together with other features such as content and genre, language and script, such clusters of literary manuscripts can contribute to scribal profiles of the anonymous scribes of the Dead Sea Scrolls, which, in turn, can shed new light on ancient Jewish scribal culture, in Hebrew and Aramaic, in the Graeco-Roman period. Here, we first tackle the palaeographic identification of these unknown scribes.
\section{Materials and methods}
In this section, we provide descriptions of:
\begin{itemize}
    \item the dataset and the image preprocessing techniques(\ref{subsec:dataset}), 
    \item the primary analysis for textural features using pattern recognition techniques, for allographic features using artificial neural networks and a combination thereof (\ref{subsec:primAnalysis}), 
    \item the second-level analysis using a different shape feature and performing statistical evaluation of the quality of the primary analysis (\ref{subsec:seconAnalysis}), and 
    \item the third-level post-hoc visual analysis (\ref{subsec:thirdAnalysis}).
\end{itemize}
Additional details and descriptions can be found in the supplementary materials.

%%%%%%%%%%%%%%%%%%%%%%%%%%%%%%%%%%%%%%%%%%%%%%%%%%%%%%%%%%%%%%%%%%%%%%%%%%%%%%%%
%%%%%%%%%%%%%%%%%%%%%%%%%%%%%%%%%%%%%%%%%%%%%%%%%%%%%%%%%%%%%%%%%%%%%%%%%%%%%%%%
%%%%%%%%%%%%%%%%% DATASET %%%%%%%%%%%%%%%%%%%%%%%%%%%%%%%%%%%%%%%%%%%%%%%%%%%%%%
\subsection{Dataset and image preparation} \label{subsec:dataset}
In this study, we have used digital images of 1QIsa\textsuperscript{a} kindly provided to us by Brill Publishers \cite{Lim1995brill}. There are $2463$ images in the Brill scrolls collection with varied resolutions from $600$ by $600$ pixels to $2800$ by $3400$ pixels, approximately. For 1QIsa\textsuperscript{a}, we have images for columns 1–54 except for columns 16 and 46 (instead, columns 15 and 47 appear twice in the Brill collection). The list of scan numbers and their corresponding column numbers are attached in the supplementary material \ref{tab:appen1:img}. For the second-level analysis, we have also used the most recent digitized multi-spectral images of the Dead Sea Scrolls, kindly provided to us by the Israel Antiquities Authority (IAA); these images are also accessible on their Leon Levy Dead Sea Scrolls Digital Library website \cite{dssllnet}. Although the IAA images do not contain any newly digitized version of 1QIsa\textsuperscript{a}, we have used this vast collection to extract dominant character shapes and produce self-organizing feature maps (see section \ref{subsec:seconAnalysis}).

The images of 1QIsa\textsuperscript{a} pass through multiple preprocessing measures to become suitable for pattern recognition-based techniques. Our first step in preprocessing is the image-binarization technique. In order to prevent any classification of the text-column images on the basis of irrelevant background patterns, a thorough binarization technique (BiNet) was applied, keeping the original ink traces intact \cite{dhali2019binet}. After performing the binarization, the images were cleaned further by removing the adjacent columns that partially appear on the target columns' images. Finally, few minor affine transformations and stretching corrections were performed in a restrictive manner. These corrections are also targeted for aligning the texts where the text lines get twisted due to the degradation of the leather writing surface (see Fig.\ref{fig:method2}). A more detailed explanation of image preparation can be found in the supplementary material \ref{appen:preprocess}.
\begin{figure}[!ht]
    \centering
    \includegraphics[width=.8\textwidth]{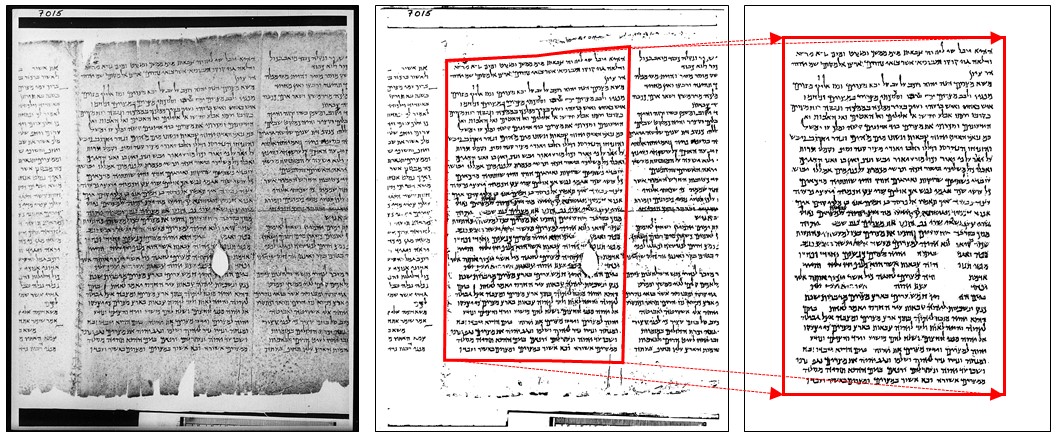}
	\caption{(\textit{from left to right}) Greyscale image of column 15, the corresponding binarized image using BiNet, and the cleaned-corrected image. From the red boxes of the last two images one can see how the rotation and the geometric transformation is corrected to yield a better image for further processing.}
	\label{fig:method2}
\end{figure}

%%%%%%%%%%%%%%%%%%%%%%%%%%%%%%%%%%%%%%%%%%%%%%%%%%%%%%%%%%%%%%%%%%%%%%%%%%%%%%%%
%%%%%%%%%%%%%%%%%%%%%%%%%%%%%%%%%%%%%%%%%%%%%%%%%%%%%%%%%%%%%%%%%%%%%%%%%%%%%%%%
%%%%%%%%%%%%%%%%% PRIMARY TESTS %%%%%%%%%%%%%%%%%%%%%%%%%%%%%%%%%%%%%%%%%%%%%%%%
\subsection{Primary analyses: Feature-space explorations} \label{subsec:primAnalysis}
In order to represent the handwriting of 1QIsa\textsuperscript{a}, we applied feature extraction methods on the binarized cleaned images to translate the handwriting style into feature vectors. The data relates directly to the tangible evidence of the ink traces in the scrolls, ink penned by scribes. As writing is a moving process that involves muscle movements of the hand and arm it is determined by the rules of physics and can therefore be quantified. 

Our feature extraction methods correlate the ink traces with the hands of the scribes on multiple levels. The allograph level of the whole character shape is easier to communicate to an audience, whereas the micro-level of textural features, such as Hinge, stands further away from the traditional visualization in the form of a palaeographic chart showing the whole character shape. Nonetheless, all these levels are equally directly related to the writing activity of ancient scribal hands that penned the ink on the scrolls. 

The question regarding 1QIsa\textsuperscript{a} whether there are different scribes or one scribe was communicated to the researcher performing the primary analysis but no further information about the state of the art regarding this question in scrolls studies (see section \ref{section:Introductio}) was communicated.

\medskip
The primary analysis involved three steps.

\medskip
\textbf{Step 1.} We have used three types of feature extraction techniques (detailed descriptions can be found in the supplementary materials \ref{appen:featureText}, \ref{appen:featureAllo}, and \ref{appen:featureAd}):
\begin{itemize}
    \item Textural feature extraction using pattern recognition techniques
    \item Allographic feature extraction using artificial neural networks
    \item Adjoined feature (a weighted combination of both textural and allographic features)
\end{itemize}

\medskip
\textbf{Step 2.} After extracting features from each of the column images, we measured the distance between the feature files using the chi-square distance. The chi-square distance $d (x,y)$ is the distance between two histograms, namely $x = [x_1,..,x_n]$ and $y = [y_1,...,y_n]$, both having $n$ number of bins. In our case, the histograms are the feature vectors. During the calculation, we normalize the histograms, i.e. their entries sum up to one. The name of the distance is derived from Pearson’s chi-square test statistics and the distance is defined as:
\begin{equation}
    d(x,y) = \frac{1}{2} \sum \frac{(x_i-y_i)^2}{x_i+y_i}
\end{equation}
These distance files contain numbers which are relatively difficult to analyze without any reference distance. To solve this issue, we first move to clustering-techniques and then to probability curves. While clustering, we reduce the feature space into a three-dimensional space to facilitate the visualization of the feature vectors. 

\medskip
\textbf{Step 3.} A feature extraction method such as Hinge provides us with a large feature vector, containing hundreds of variables. Some features in the feature vector might not have a large influence on the result. Therefore, the dimensionality of the data can be reduced in such a way that the most important aspects of the data remain. One way to do this is using Principal Component Analysis (PCA). It transforms the data into \textit{n} components that are independent of each other. Using PCA we go from multidimensionality to a three-dimensional space, and then inspect this three-dimensional plot to see if there is any significant movement of the point cloud. 

\medskip
\noindent In order to facilitate the decision-making process directly from the distance files (from step 2), one typical approach is to analyse probability curves; a False Acceptance Rate (FAR) curve (the likelihood that the system will incorrectly \textit{accept} a writer) and a False Reject Rate (FRR) curve (the likelihood that the system will incorrectly \textit{reject} a writer). These curves are generated from a known set of writers to incorporate all the variabilities. Depending on the distance between two feature vectors, the probability of being the same or a different writer can be determined. Unfortunately, in the Dead Sea Scrolls collection, there is \textit{no certain} identification of known writers. In this study, we have avoided to introduce into our algorithm any \textit{assumptions} by palaeographers about scribal identity or difference in the scrolls in general or in 1QIsa\textsuperscript{a} specifically. This procedure ensures the outcome of this study to be independent from any bias. 

Instead of being able to use probability curves, robust alternative techniques are needed for the Dead Sea Scrolls. In order to cross-check and test the quality of our findings from the primary analysis, we have used statistical evaluation as second-level analysis.

%%%%%%%%%%%%%%%%%%%%%%%%%%%%%%%%%%%%%%%%%%%%%%%%%%%%%%%%%%%%%%%%%%%%%%%%%%%%%%%%
%%%%%%%%%%%%%%%%%%%%%%%%%%%%%%%%%%%%%%%%%%%%%%%%%%%%%%%%%%%%%%%%%%%%%%%%%%%%%%%%
%%%%%%%%%%%%%%%%% SECONDARY TESTS  %%%%%%%%%%%%%%%%%%%%%%%%%%%%%%%%%%%%%%%%%%%%%
\subsection{Secondary analyses: Statistical evaluations} \label{subsec:seconAnalysis}
The second-level analyses' goal is to independently assess whether there is a transition of style in the sequence of columns. The suspicion that there is a transition in the series of columns was communicated to the researcher performing this cross-check. However, until step 5, no more specific information was given about the sequence of columns where a style transition was observed in the primary analysis. The logistic tests performed in this part of the study were not influenced by any column information. This procedure ensures the independence of the second-level cross-examination. 

\medskip
The second-level analysis involved five steps; more detailed descriptions can be found in the supplementary materials \ref{appen3:sec:second}.

\medskip
\textbf{Step 1.} In order to use a shape feature that is very different from those used in the primary phase of the study, it was proposed to use a fraglet approach, the so-called fragmented-connected component contours (fco3) \cite{schomaker2004automatic,bulacu2007text,schomaker2007using}. In comparison to textural features that are concentrated around micro-details along the ink trace, fraglets contain more allographic information that may be understandable to a paleographer.

\medskip
\textbf{Step 2.} A large Kohonen self-organizing feature map (SOFM) was computed, containing $80\times80$ centroids for such fraglets from the total IAA multi-spectral images collection that is at our disposal, yielding $6400$ prototypical fraglets. About $600k$ randomly selected fraglets were used for this stage. Each centroid is based on about $94$ fraglet instances. The use of the Kohonen map is not essential. Other clustering methods can be used; this step is not critical. But the Kohonen map has the advantage that the centroids that end up in the map change gradually, as opposed to a haphazard result of the ordering of centroids in, e.g., a k-means algorithm. 

\medskip
\textbf{Step 3.} For the series of columns, a histogram was computed for split-scan samples $a$ and $b$, separately. In digital paleography and forensic handwriting this approach is used in order to check a reasonable response of the algorithm. It is expected that version $a$ and $b$ of a column of text should be close neighbours, under the assumption that a column was produced by a single scribe. If a hit list of neighbours for a query $a$ of a column does not return the corresponding $b$ version in the top of the hit list of a search operation, results should be judged critically. Conversely, if the corresponding sample $b$ appears at the top, the neighbouring hits will also have a larger probability of being produced by the same scribe~\cite{GIWIS}.

\medskip
\textbf{Step 4.} For each sample, the nearest neighbours were computed in the rest of the list. Bookkeeping was performed on the distance in feature space and the column number of the hits that were found.

\medskip
\textbf{Step 5.} From the computed data (from steps 1-4), i.e., the distances and the column numbers of the nearest neighbour samples, four follow-up steps can be taken that help to determine whether the handwriting style is uniform throughout the manuscript of 1QIsa\textsuperscript{a} or whether there are style differences.

\textbf{5a.} For testing the deviation of a random {\em voting pattern} for left-vs-right neighbours of a given column in fraglet-shape space, a Chi-square test was used. If there is a single signal source (scribe), nearest neighbors will fall to the left or right of a column in the series in a random pattern.

\textbf{5b.} A one-way analysis of variance, a t-test, was performed on the {\em distance values} of the left versus right nearest-neighbour matches in the series of columns.

\textbf{5c.} Apart from the distance between columns in fraglet shape space, it is interesting to check the {\em estimated position} of a best-matching neighbour column for any given column in the series. If there is a single scribe, the nearest neighbour would appear in any column in the scroll. Conversely, if there are two scribes, the columns on the left would tend to have their best-matching neighbours on the left, and vice versa.

\textbf{5d.} If there is a phase transition in the sequence of columns, fitting a logistic curve on the variable `average neighbour position' over columns should reveal the switching point reliably, i.e., with a high Pearson correlation of the fit. The number of the critical phase-transition column is the output of this test.

%%%%%%%%%%%%%%%%%%%%%%%%%%%%%%%%%%%%%%%%%%%%%%%%%%%%%%%%%%%%%%%%%%%%%%%%%%%%%%%%
%%%%%%%%%%%%%%%%%%%%%%%%%%%%%%%%%%%%%%%%%%%%%%%%%%%%%%%%%%%%%%%%%%%%%%%%%%%%%%%%
%%%%%%%%%%%%%%%%% TERTIARY TESTS  %%%%%%%%%%%%%%%%%%%%%%%%%%%%%%%%%%%%%%%%%%%%%%
\subsection{Tertiary analyses: Post-hoc visual analyses} \label{subsec:thirdAnalysis}
The aim of the post-hoc visual analyses was to attempt to correlate the quantitative analyses from pattern recognition and artificial intelligence techniques with a qualitative analysis from a traditional palaeographic approach. 

\medskip
The third-level analysis involved three steps.

\medskip
\textbf{Step 1.} For visual inspection by palaeographers, we created charts with full character shapes for individual Hebrew letters that can be found in the supplementary material \ref{appensec:charts}. 

\medskip
\textbf{Step 2.} In order to facilitate the complex process of visual inspection, we generated heatmaps for each character shape. The heatmaps are aggregated visualizations of the shape of each letter. These are made up of all particular instances of a letter and as such do not exist in one particular form. Thus, the use of heatmaps fulfils, through a sophisticated and robust procedure, the requirement from forensics to study each particular instance of a character. Also, the visualization by heatmaps may be an important step forward because they could work better than the palaeographic charts used traditionally in the field as they are not limited to one or more particular examples of which the indicative value can be doubtful but are made up of all instances of a letter. 

\medskip
\textbf{Step 3.} Suppose the primary analyses' results and the statistical tests in the second stage would turn out significant. In that case, a post-hoc visual analysis of the fraglet set contributing best to the discrimination between the left and right parts of the sequence is required to bridge the quantitative and the qualitative approaches. The fraglets refer to the characters' parts that can be more precise, distinctive, and informative in finding significant shape differences than the full characters. For each of the fraglet shapes, exploration can be performed to identify their significance in separating two halves (if there exists a separation). Then, by running all possible combinations of $6400$ fraglets and counting their presence in each image,  a statistical view of the two halves can be obtained. 
\section{Results}
\subsection{Primary analyses} \label{subsec:first}
\begin{figure}[!ht]
    \centering
    \includegraphics[width=.98\textwidth]{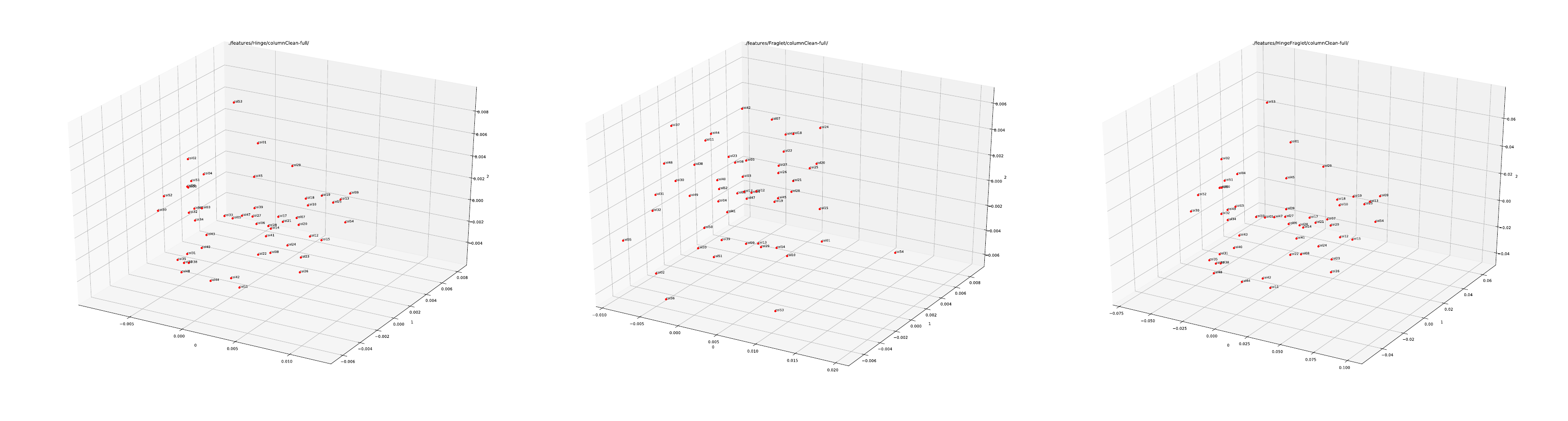}
	\caption{Plots of different feature vectors in three-dimensional space using PCA (\textit{from left to write}: Hinge, Fraglet, and Adjoined features).}
	\label{fig:res:allred}
\end{figure}

Here we present the plots that result from the three types of feature extraction techniques that we used and the distance measurements between the feature files using the chi-square distance. The plots have been examined to find any possible clustering or any significant movement of the points' cloud in the columns of 1QIsa\textsuperscript{a}. We used the PCA technique on each of the feature collections and plotted them in a three-dimensional visual space (see Fig.\ref{fig:res:allred}). Figure \ref{fig:res:allred} shows the red points for each of the columns of 1QIsa\textsuperscript{a}.  

\begin{figure}[!ht]
    \centering
    \includegraphics[width=.6\textwidth]{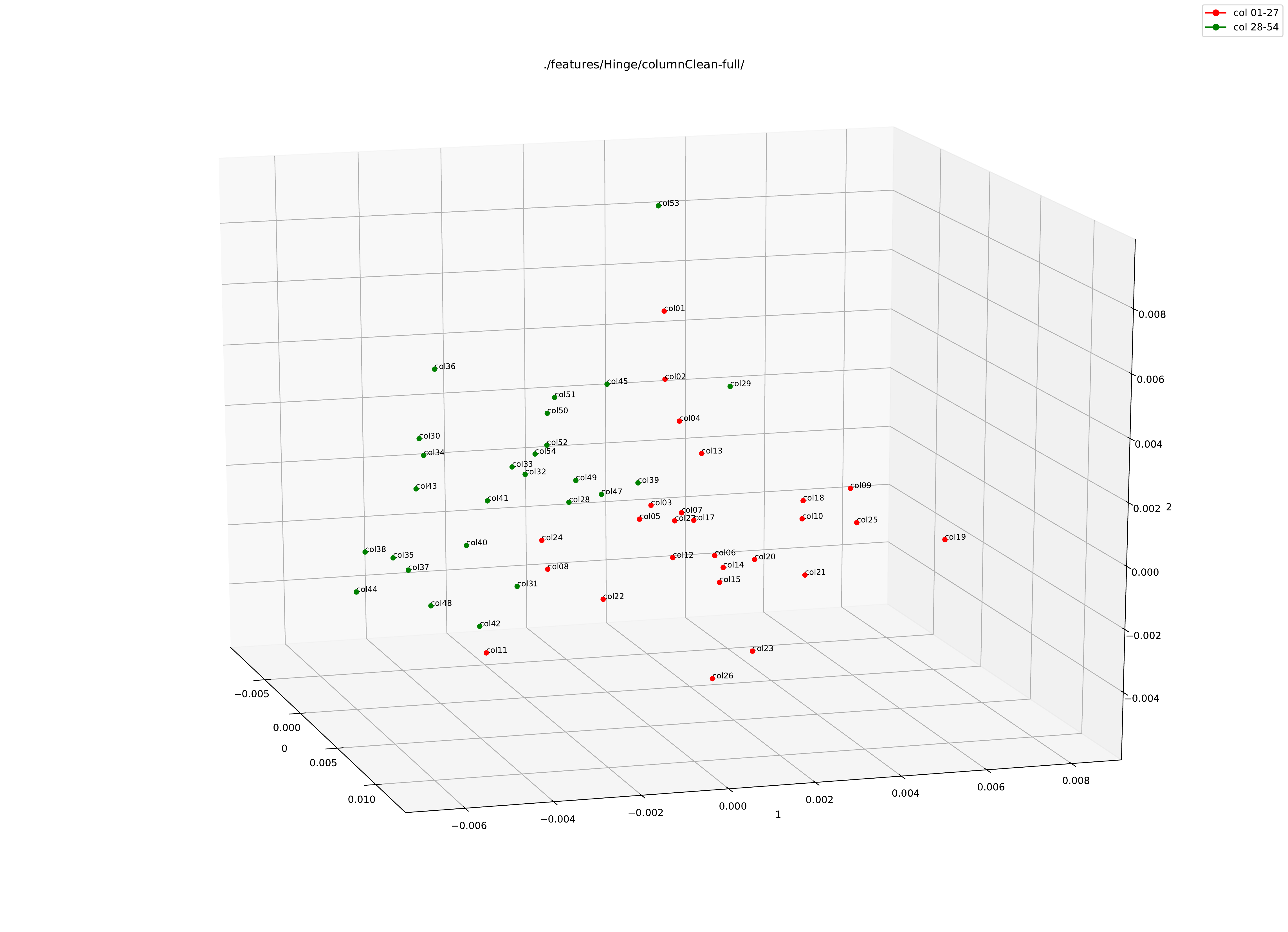}
	\caption{Hinge feature plot (using PCA) of the full column images of 1QIsa\textsuperscript{a} (red: columns 1–27; green: column 28–54)}
	\label{fig:res:hinge-full}
\end{figure}

\begin{figure}[!ht]
    \centering
    \includegraphics[width=.6\textwidth]{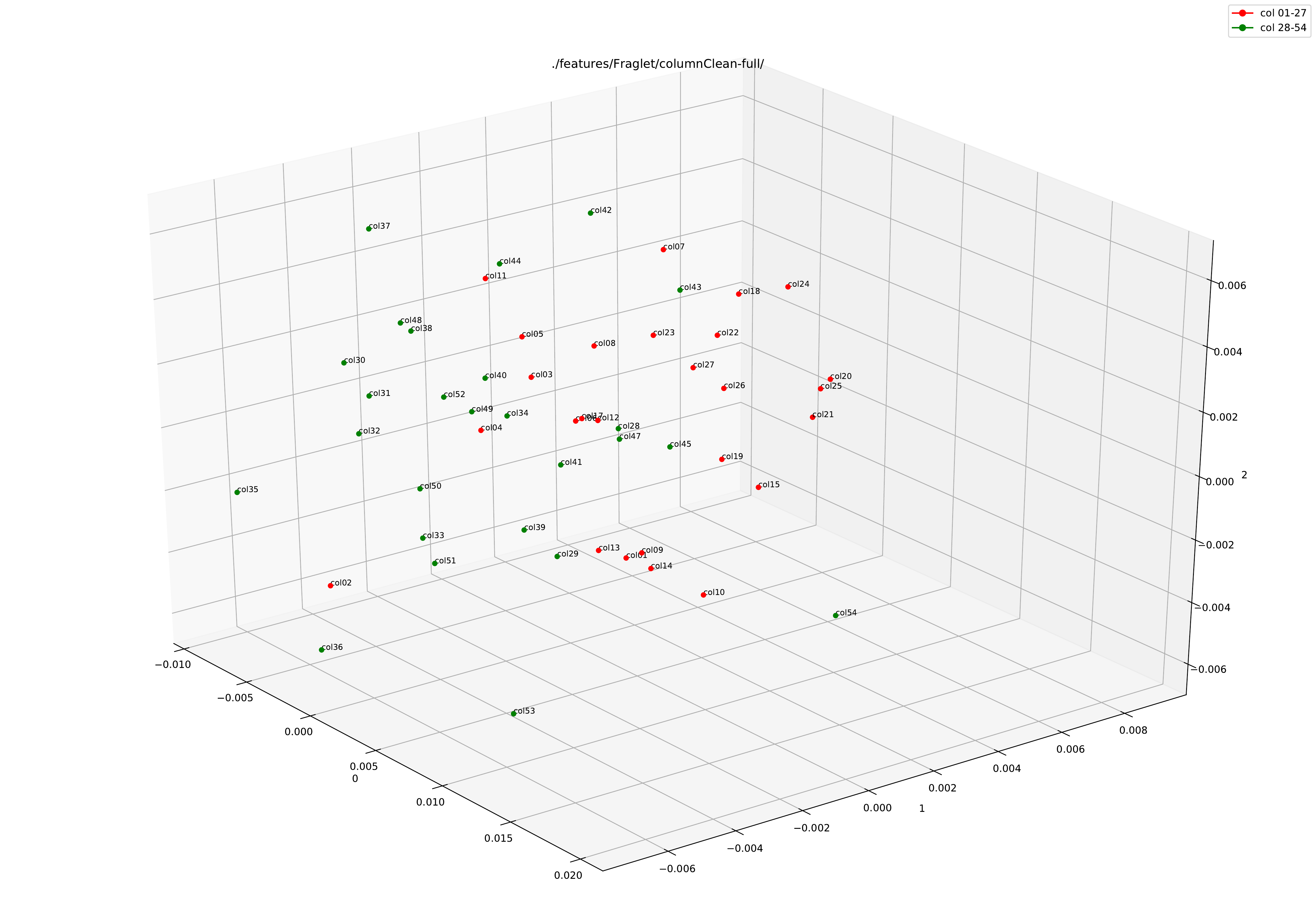}
	\caption{Fraglet feature plot (using PCA) of the full column images of 1QIsa\textsuperscript{a} (red: columns 1–27; green: column 28–54).}
	\label{fig:res:fraglet-full}
\end{figure}

\begin{figure}[!ht]
    \centering
    \includegraphics[width=.65\textwidth]{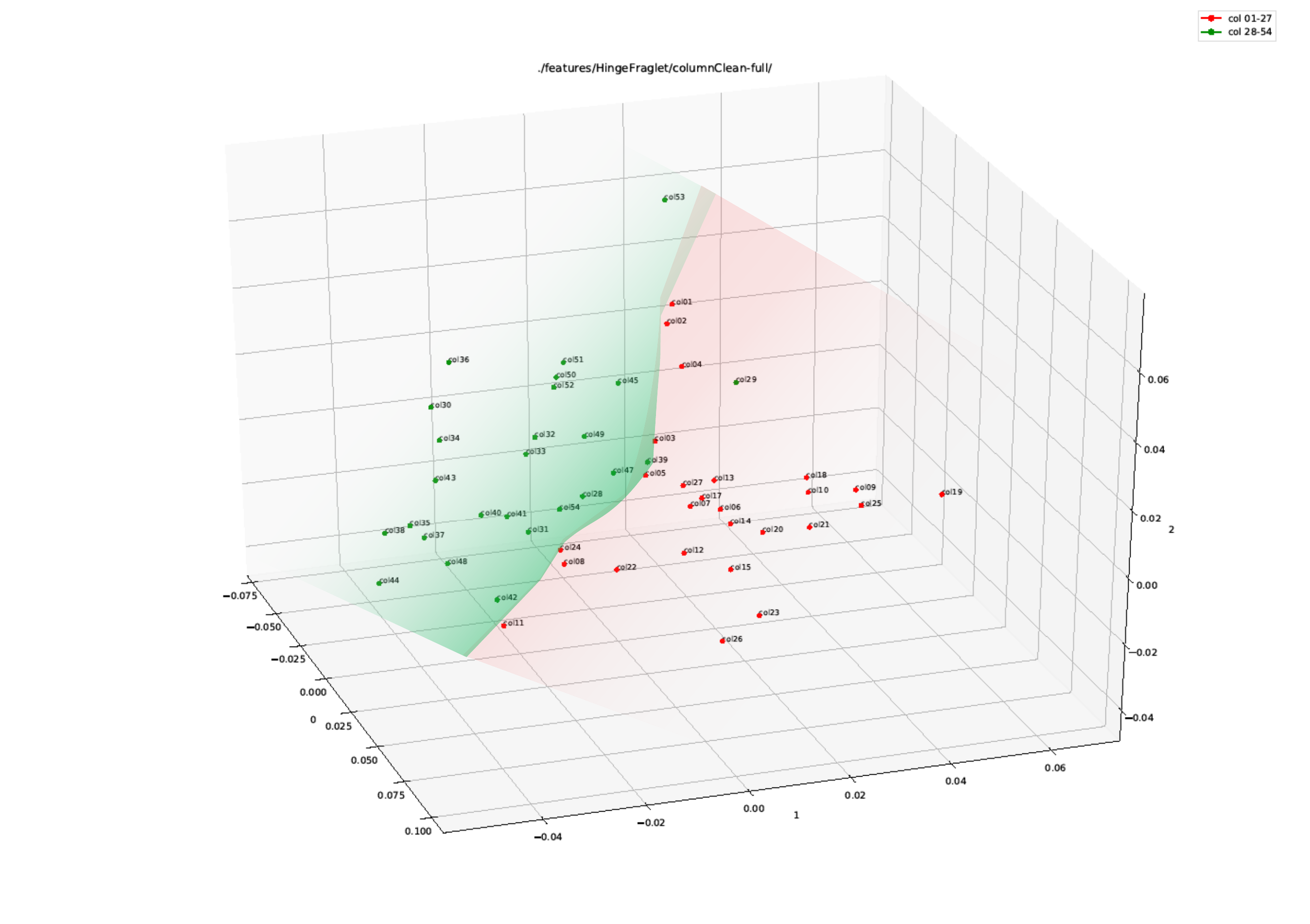}
	\caption{Adjoined (Hinge+Fraglet) feature plot (using PCA) of the full column images of 1QIsa\textsuperscript{a} (red: columns 1–27; green: column 28–54).}
	\label{fig:res:Hinge-fraglet-full}
\end{figure}

In the next step, we used the colour red for columns 1–27 and the colour green for columns 28–54. Please note that this colouring works just as a label and has no effect/consequence on the experiments. The plots were then generated again for all the three types of features (see Fig.\ref{fig:res:hinge-full}, Fig.\ref{fig:res:fraglet-full}, and Fig.\ref{fig:res:Hinge-fraglet-full}).

\begin{figure}[!ht]
    \centering
    \includegraphics[width=.65\textwidth]{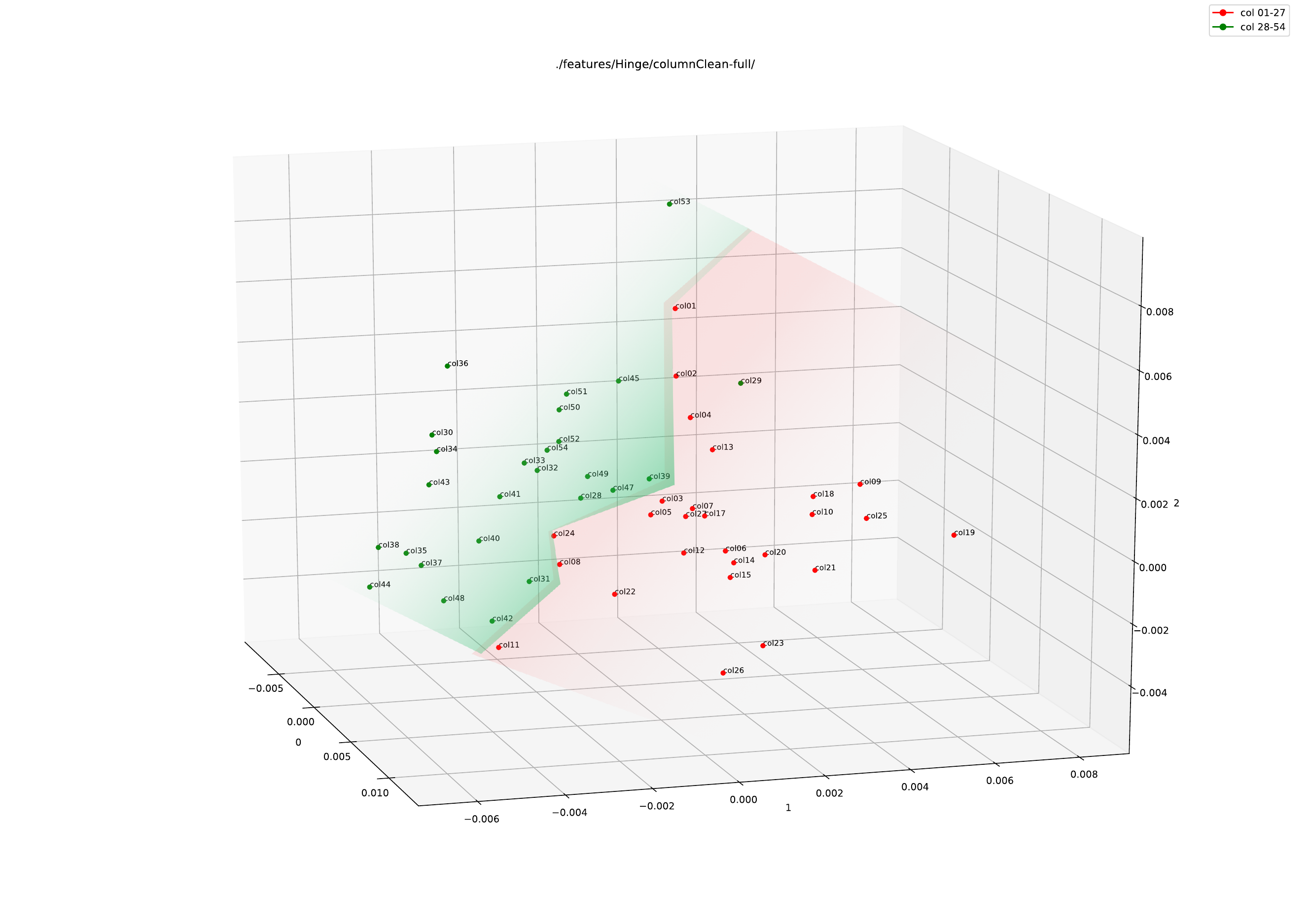}
	\caption{Hinge feature plot (using PCA) of the full column images of 1QIsa\textsuperscript{a} with a clear line of separation (red: columns 1–27; green: column 28–54).}
	\label{fig:res:hinge-full-color}
\end{figure}

Figure \ref{fig:res:hinge-full} shows the plot using Hinge feature vectors on the full column images of 1QIsa\textsuperscript{a}. There is a separation between the two sets of columns. Except for an outlier (column 29), the red and green points can be separated using a two-dimensional plane (similar to a piece of paper). This is visualized in Figure \ref{fig:res:hinge-full-color}. The implication is that there might be a clear separation of the two sets of data, yet they are also close to each other. 

As for column 29 appearing as an outlier in this part of the primary analysis, in the independent second-level analyses (see section \ref{subsec:second}), column 29 does not show up as a clear outlier. Also, in the primary analysis, column 29 is not an extreme outlier. Instead, it is close to the separation line of the two halves of the manuscript. Further tests can be performed in the future to conclude on a concrete reason for this.

Figure \ref{fig:res:fraglet-full} shows the plot for the Fraglet feature from a $70\times70$ Kohonen SOFM. Here, the points are not that clearly separated as in the case of the Hinge feature. The reason might be because the Fraglet feature renders the physical shapes of characters, similar to what the human eyes see, it is less adequate (in this particular case) to determine any micro-level differences in the data. 

In the last step, we combined both these features, Hinge and Fraglet. Figure \ref{fig:res:Hinge-fraglet-full} shows the plot for the combined feature, or Adjoined feature. A clear separation is visible here between the data points in the adjoined feature plot.

Thus, the primary analysis indicates a significant difference between the two halves of the columns of 1QIsa\textsuperscript{a} with a visibly clear separation in the points' cloud of features.

%%%%%%%%%%%%%%%%%%%%%%%%%%%%%%%%%%%%%%%%%%%%%%%%%%%%%%%%%%%%%%%%%%%%%%%%%%%%%%%%
%%%%%%%%%%%%%%%%%%%%%%%%%%%%%%%%%%%%%%%%%%%%%%%%%%%%%%%%%%%%%%%%%%%%%%%%%%%%%%%%
%%%%%%%%%%%%%%%%% SECONDARY ANALYSES  %%%%%%%%%%%%%%%%%%%%%%%%%%%%%%%%%%%%%%%%%%
\subsection{Secondary analyses} \label{subsec:second}
Steps 1–4 (described in section \ref{subsec:seconAnalysis}) are pre-requisites to perform the tests in step 5. A detailed description of the first four steps can be found in the supplementary material \ref{appen3:kohonen:80}. It is important to note here that the fraglet features (fco3) used in the secondary analyses are derived from a different Kohonen SOFM than those used in the primary analyses (supplementary materials \ref{appen:featureAllo} and \ref{appen3:kohonen:80}). This is done to ensure the independence of two analyses and to perform cross-validation. The results of the statistical tests conducted in step 5 of second-level analyses are as follows.

\begin{figure}[!ht]
    \centering
    \includegraphics[width=.45\textwidth]{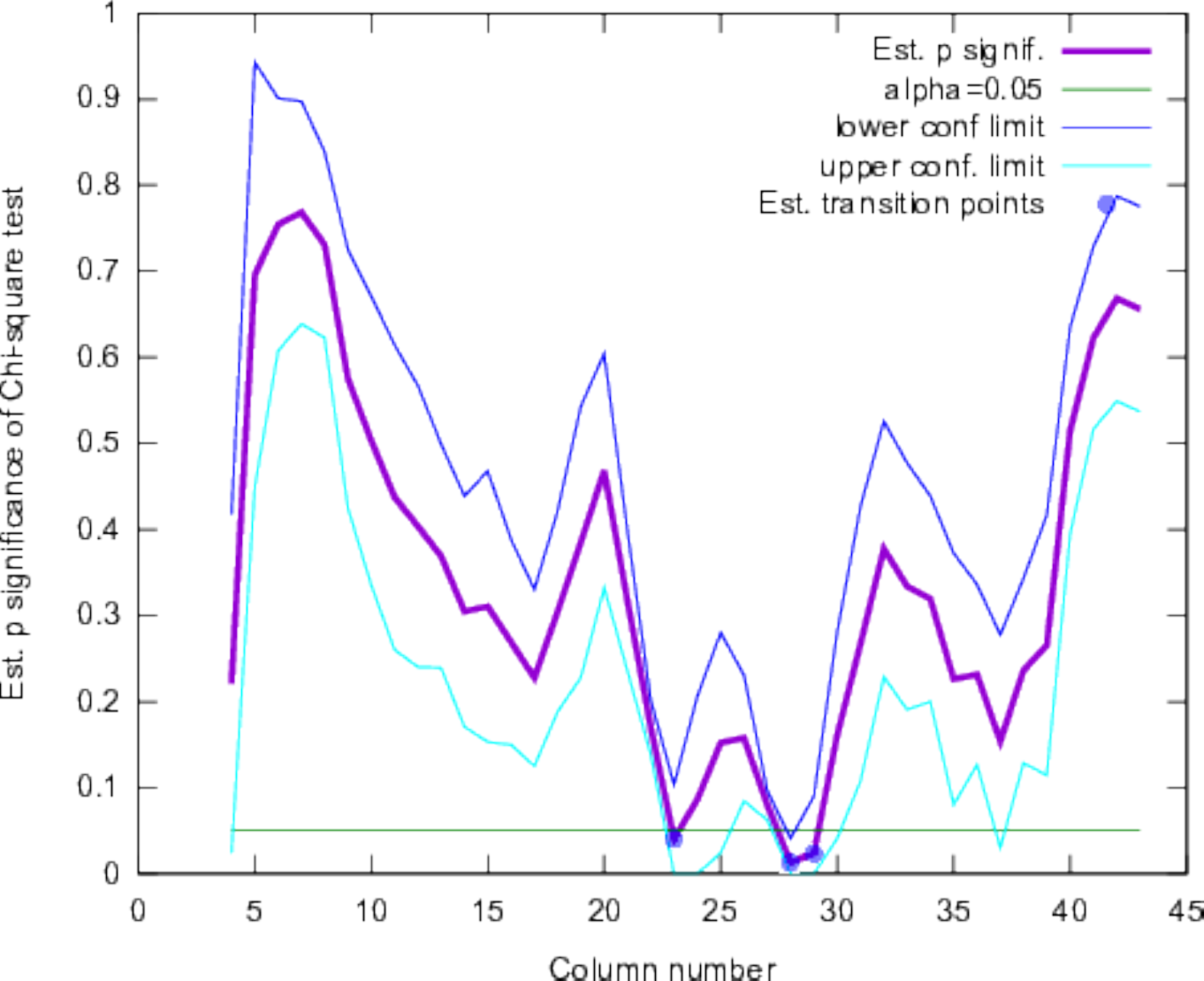}
    \caption{ Estimated significance probability for a left vs right Chi-square test, averaged over a range of window sizes of 9-26. With an alpha of 0.05, columns 23, 28, and 29 indicate that there is a discrepancy between the left and right neighbour votes. The general pattern suggests that something is changing in the statistics of the hit pattern left/right, around the middle of the column sequence.}
    \label{fig:chisquarepattern5a}
\end{figure}

\medskip
\textbf{Step 5a.} Figure~\ref{fig:chisquarepattern5a} shows the pattern of statistical probability that the left/right voting pattern deviates from random. A clear dip is present at the middle of the graph, confirming that at that point, the probability of nearest neighbours of a column falling to the left or right is very likely not an accident. This analysis, however, would be considered exploratory, and not a rigorous test, due to multiple testing over several time windows. Therefore, additional testing was done on the basis of the pattern of distances of columns to their nearest neighbours in shape space.

\medskip
\textbf{Step 5b.} The average distance from query column to best match is $0.238$ ($sd=0.003$) on the left vs $0.231$ ($sd=0.008$) on the right, $p=0.002$, which is significant at $\alpha=0.005$. The inter-column distances are somewhat higher in the left series as compared to the right part, but their regularity is higher, given the lower standard deviation.

\begin{figure}[!ht]
    \centering
    \includegraphics[width=.5\textwidth]{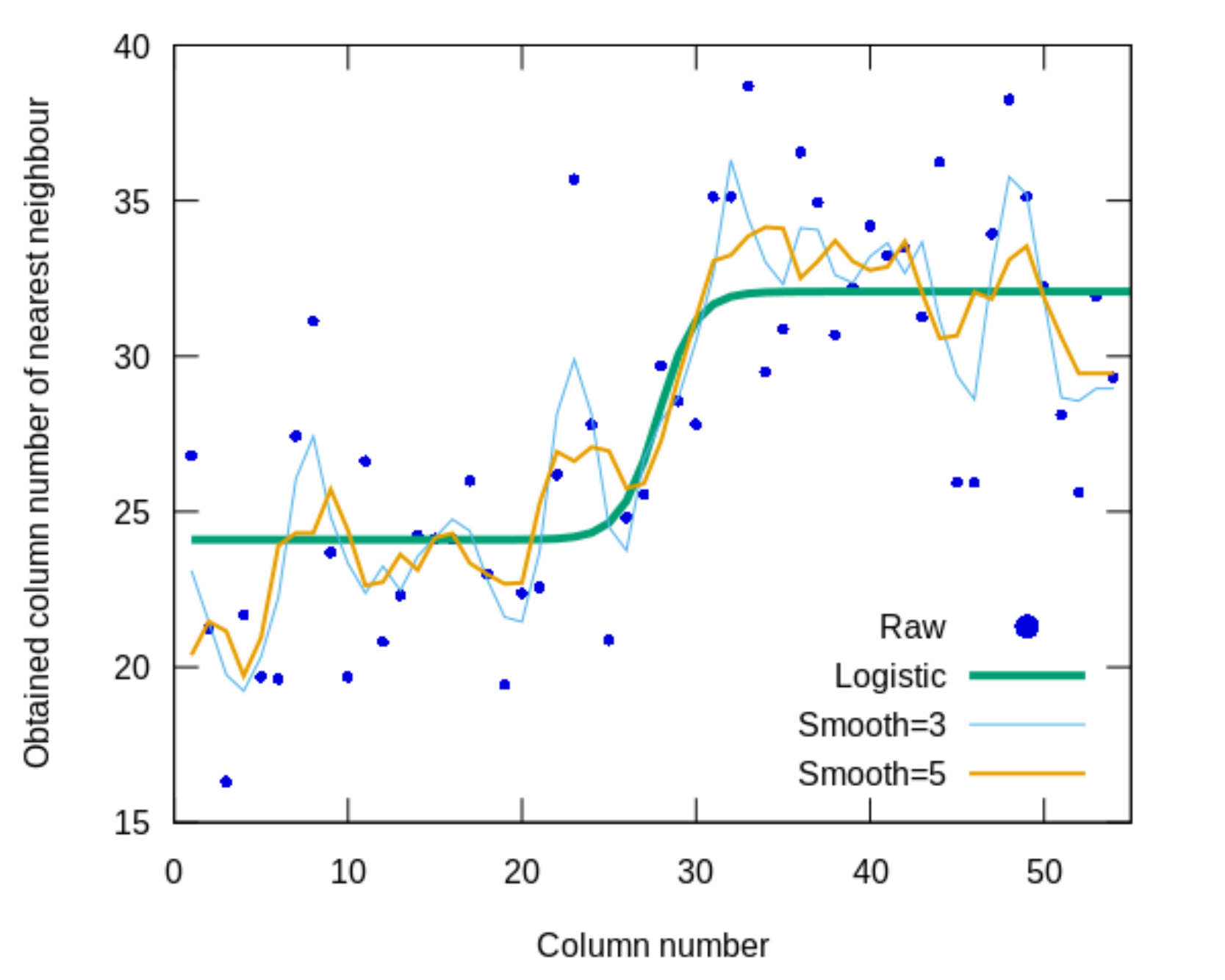}
    \caption{Average serial column position of the nearest-neighbour of a column, measured in fraglet-feature space. Raw samples (points), curves smoothed over 3 and over 5 points, and a best-fit logistic curve}
    \label{fig:fitlogistic}
\end{figure}

\medskip
\textbf{Step 5c.} Figure~\ref{fig:fitlogistic} shows the obtained column position of the best fitting neighbour for a column. Visually, from the smoothed curves, it can be seen that left of column 27, the average position of the hits is between column 20 through 25. On the right of column 27, the average position of hits is between column 30 and 35. A t-test indicates that the average nearest-neighbour column number for a column on the left is at column 24 ($sd=4$), for a column on the right it is position 32 ($sd=3.7$), where $p < 0.0001$ (see supplementary material \ref{appen3:5b}). Therefore the between-column similarity is highest 'ipsilateral' with respect to the cut point (column 27): '\textit{left}' looks like left, '\textit{right}' looks like right.

\medskip
\textbf{Step 5d.} The results from steps 5a-5c are visually and statistically clear, but the valid question may be asked whether the actual point, i.e., the {\em column number of a phase transition} can also be computed? The logistic function or Fermi-Dirac function is usually used to model phase transitions in physics and biology. In the humanities, it can be used to model language change~\cite{bod2003probabilistic,kroch1989reflexes}. Two types of analysis were performed to estimate the parameters of the logistic model:
\begin{equation}
    f(x) = Y_{offset} + \frac{A}{1+exp(b(x-x_{offset})}
\end{equation}

\noindent where $x$ is the column number, $f(x)$ is the estimated average position of its nearest neighbour in the column series as measured in the fraglet shape space. Parameter $Y_{offset}$ represents the vertical offset, $A$ represents the scale factor, $b$ represents the steepness of the phase transition and $x_{offset}$ represents the column number where the phase transition occurs.In order to be very sure that a solution for the transition point is not haphazard, we will perform two very different estimation procedures for the logistic function. First, in order to allow a list of high-quality model fits to evolve without constraints, we used a Monte-Carlo estimation, randomly varying parameter values and remembering the best solutions. This sampling approach allows good results to emerge, without theoretical assumption. The second method is the more traditional curve-fitting approach that uses the 'least-squares error' as the assumed constraint, to deliver a single best-effort solution. Without seeding a logistic function estimator with knowledge concerning the suspected column number 27, the output of the Monte-Carlo estimate can be found in Table \ref{tab:logisticpar}:
\begin{table}[!ht]
    \centering
    \caption{Resulting parameter values for a Monte-Carlo estimated logistic model ($r=074$)}
    \begin{tabular}{l|l}
    \hline
    parameter & value \\ \hline  \hline
    $x_{offset}$ &  27.8 \\
    $Y_{offset}$ &  24.1 \\
    $A$          &  7.98 \\
    b            &  0.92 \\ \hline
    \end{tabular}
    \label{tab:logisticpar}
\end{table}

\noindent In Table \ref{tab:logisticpar}, the value of $x_{offset}$ means that the transition column is estimated to occur between column 27 and 28, with a transition steepness that is smooth lasting from column 24 to 32 (Fig~\ref{fig:fitlogistic}). The fit of the sigmoid transition model is significant, with a correlation r that equals 0.74 ($p < 0.001$) on the raw data. An exact fit would have yielded r=1.0. Although not perfect, r=0.74 would be considered as a very robust correlation in empirical disciplines such as psychology and biology. The model would explain 55\% of the variance in the data, which is not strange, given the fact that the logistic model is a stylized description of a time sequence with irregularities. If we smooth the irregularities over time, using a running average over a limited 3 or over 5 samples (columns), to smooth out the within writer variation, the correlation with the sigmoid increases considerably: if we smooth the column time series over 3 values, $r=0.87$ (76\% var. explained variance by sigmoid phase transition); if we smooth the column time series over 5 values, r=0.93 (86\% var. explained variance by sigmoid phase transition).

As a double check, the Monte Carlo-based fit was replicated with a more traditional least-squares curve fit (Python scipy package), yielding a phase transition at column 26.6 for raw data, with $r=0.74$, at 26.2 for a smoothed time series with a window of three points ($r=0.87$) and a transition at column 26 for a smoothed time series with a window of five points ($r=0.94$). This double-check indicates that both traditional curve fitting and stochastic model fits yield a transition around the middle columns of 1QIsa\textsuperscript{a}. Interestingly, also the quality of fit (correlation) is similar for these two very different estimation methods, adding to the trust in the found transition point.

\medskip
Thus, the second-level analyses confirm the presence of two different clusters in writing style in a series of handwritten columns, to be called {\em left} and {\em right}. The confirmation occurs in three different ways: 
\begin{itemize}
    \item left/right votes for the relative serial position of the nearest neighbour of a column, 
    \item distance to nearest neighbours on the left or right, and 
    \item average serial column position of the nearest neighbour of a given column in fraglet-feature space.
\end{itemize}
The results from these analyses show that a transition point occurs at around column 27, although the obtained logistic model fit in step 5d suggests that the transition may not be completely sharp.

%%%%%%%%%%%%%%%%%%%%%%%%%%%%%%%%%%%%%%%%%%%%%%%%%%%%%%%%%%%%%%%%%%%%%%%%%%%%%%%%
%%%%%%%%%%%%%%%%%%%%%%%%%%%%%%%%%%%%%%%%%%%%%%%%%%%%%%%%%%%%%%%%%%%%%%%%%%%%%%%%
%%%%%%%%%%%%%%%%% TERTIARY TESTS  %%%%%%%%%%%%%%%%%%%%%%%%%%%%%%%%%%%%%%%%%%%%%%
\subsection{Tertiary analyses}
The results for our attempt to correlate by visualization the quantitative analyses from pattern recognition and artificial intelligence techniques to the level suitable for palaeographers to be able to see what the quantitative analyses `{see}', in this case a clear separation in style, are as follows.

\medskip
\textbf{Step 1.} Our charts with full character shapes for individual Hebrew letters improve significantly on the traditional palaeographic chart, such as in \cite{ulrich2011discoveries}. Each instance of a character can be directly traced back to its exact position in the manuscript of 1QIsa\textsuperscript{a}. Also, there is no modern human hand involved, either in retracing the characters or in character reconstruction. The ink traces are extracted as is from the digital images and retain the movements once made by the ancient scribe's hand (see supplementary material \ref{appensec:charts}).

However, as described in Section \ref{section:Introductio}, due to the large number of characters from each column and the number of columns, the decision-making process from visual inspection alone of such charts may prove inadequate. 

\begin{figure}[!ht]
    \centering
    \includegraphics[width=.9\textwidth]{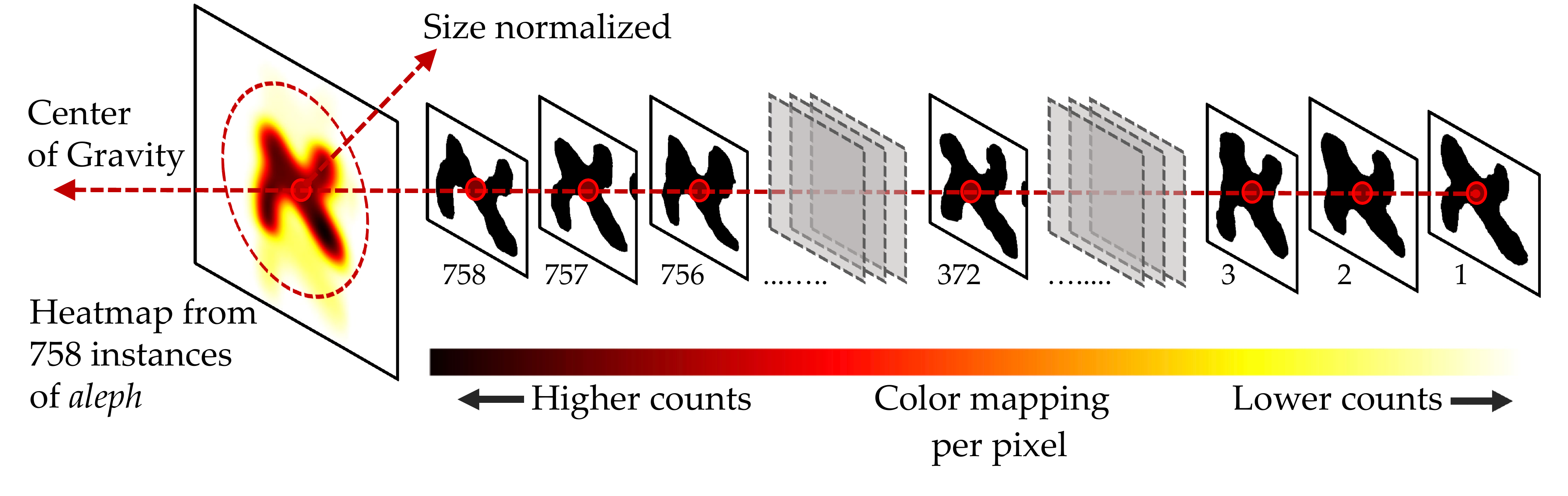}
	\caption{An illustration of how heatmaps of normalized average character-shapes are generated for individual letters (example: \textit{aleph}).}
	\label{fig:res:heatmap-generation}
\end{figure}

\medskip
\textbf{Step 2.} A character heatmap is the normalized average character shape of individual letters extracted from the column images and aligned on their centroids (see Figure \ref{fig:res:heatmap-generation}). The heatmaps are neither dependent nor produced from the primary and secondary analyses (subsections \ref{subsec:first} and \ref{subsec:second}). They are entirely independent of pattern recognition and artificial intelligence-based tests. We present these heatmaps to produce an easy-to-use visualization for the palaeographers to observe any differences between letters coming from different columns.  

\begin{figure}[!ht]
    \centering
    \includegraphics[width=\textwidth]{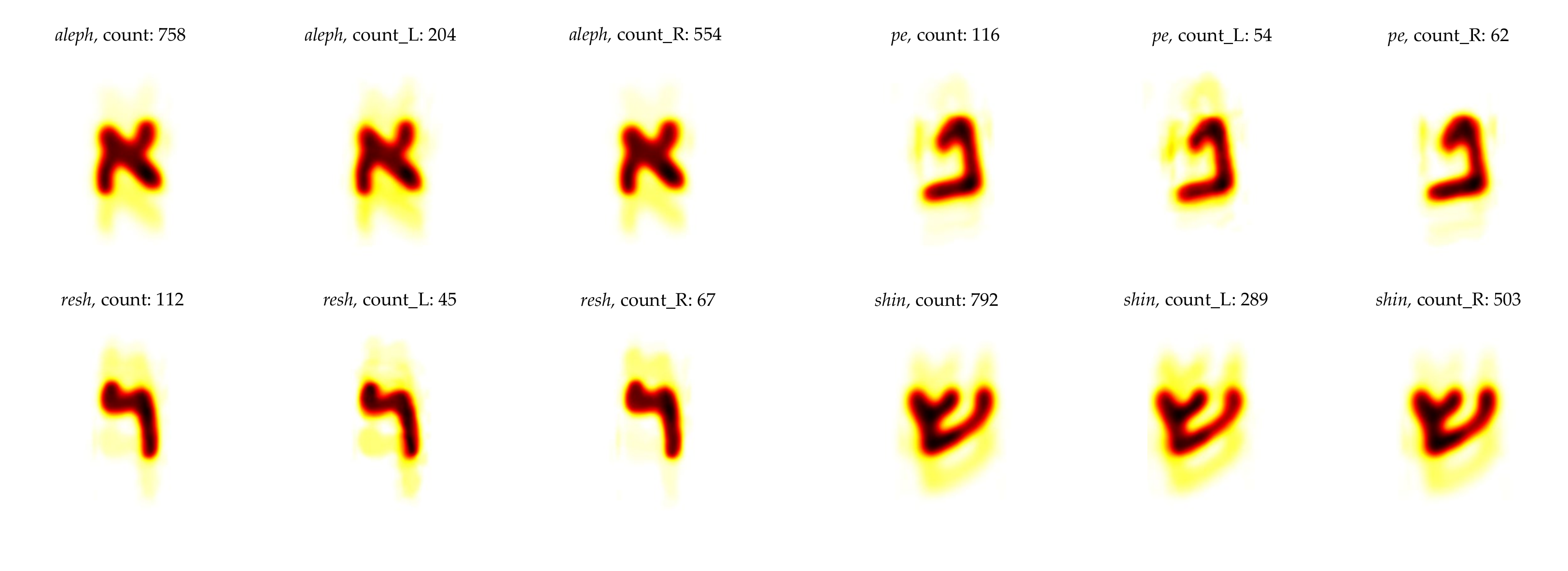}
	\caption{Individual character heatmaps of \textit{aleph}, \textit{pe}, \textit{resh}, and \textit{shin} from 1QIsa\textsuperscript{a}. On the top left, the first \textit{aleph} is aggregated from all the columns, the next one is from columns 1–27, and the final one is from columns 28–54. The same applies for the other three characters.}
	\label{fig:res:heatmaps}
\end{figure}

\begin{figure}[!ht]
    \centering
    \includegraphics[width=.98\textwidth]{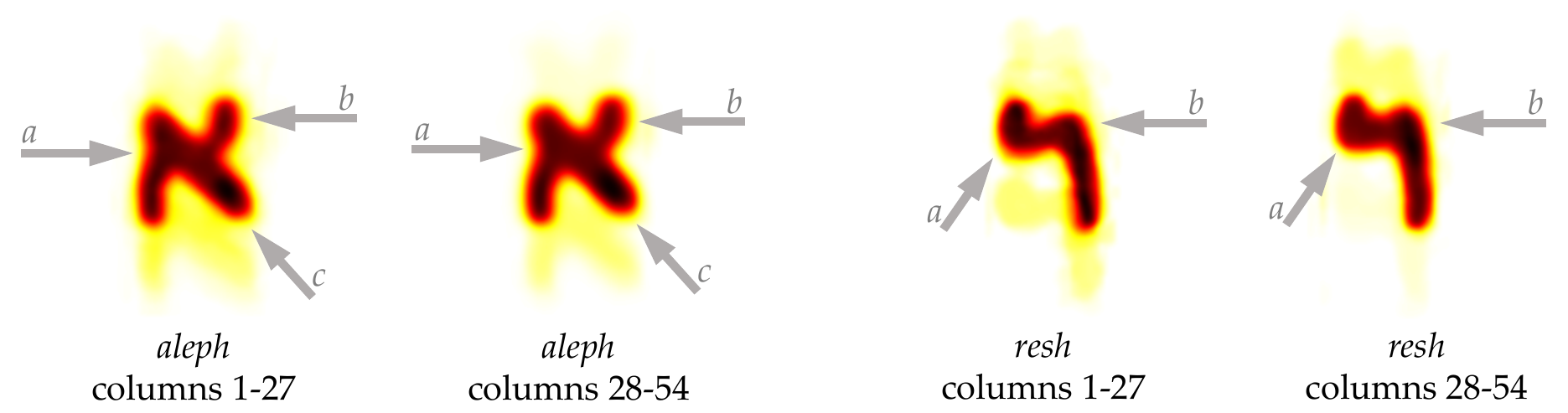}
	\caption{A zoomed in view of \textit{aleph} and \textit{resh} from Figure \ref{fig:res:heatmaps}. In the case of \textit{aleph}: the subtle difference in positioning of the left down stroke (\textit{a}), the right upper stroke vis-à-vis the diagonal stroke (\textit{b}),  and the slight difference in thickness of the diagonal stroke (\textit{c}). in the case of \textit{resh}: the curvature of the top stroke (\textit{a}), and the slight difference in thickness and length of the horizontal stroke (\textit{b}).}
	\label{fig:res:heatmap-zoomed}
\end{figure}

We generated three different heatmaps for each letter, corresponding to the three aggregate levels for all columns of 1QIsa\textsuperscript{a}, for columns 1–27, and for columns 28–54 (for some examples, see Figure \ref{fig:res:heatmaps}). Though the full-character shapes from Figure \ref{fig:res:heatmaps} seem to exhibit not that much differences among them, a close inspection reveals subtle differences between the two halves of 1QIsa\textsuperscript{a}. These differences can be observed in the thickness of strokes and the positioning of connections between strokes. See, for example, the subtle difference in positioning of the left down stroke and the right upper stroke vis-à-vis the diagonal stroke of \textit{aleph} and the slight difference in thickness of the diagonal stroke, or the slight difference in thickness and length of the horizontal stroke of \textit{resh} (see Figure \ref{fig:res:heatmap-zoomed}). 

In a traditional palaeographic chart such differences might be deemed insignificant and explicable as normal variations within the handwriting of one writer. If that were the case, i.e., what we see is normal within writer variability, then for 1QIsa\textsuperscript{a} one would expect the same distribution of writing style across all columns, which is not the case. Rather, the primary analyses as well as the statistical tests (5a–5d) indicated a significant separation and a clear distribution of the two halves of the manuscript of 1QIsa\textsuperscript{a} on either side of the divide. 

Heatmaps should be inspected with a different understanding. Heatmaps are different from traditional palaeographic charts in that they represent the aggregated visualizations of the shape of each letter, hundreds per letter in the case of 1QIsa\textsuperscript{a}. Given the large number (count) of samples and the fact that the center position estimate is stable, then the remaining differences after averaging are an indication of an underlying structural difference. Thus, in heatmaps, the subtle differences we see between the different aggregate levels are indicative \textit{if} the separation between the different levels has also turned out significant otherwise, which is the case for 1QIsa\textsuperscript{a}.

Note, that we have only used the automatically recognized characters to generate the heatmaps from the columns of 1QIsa\textsuperscript{a}. The number of generated \textit{aleph}s for the heatmaps is 758, while the total number of \textit{aleph}s in 1QIsa\textsuperscript{a} is 5011. These 758 \textit{aleph}s were automatically extracted by the computer on the basis of known shape structures, and the extracted characters come from all columns, representing a general distribution. This extraction is extremely efficient and has the advantage that it does not require human intervention. Our goal is not to produce an exhaustive enumeration of all \textit{aleph}s in the manuscript, rather than produce heatmaps that cover all columns with a sufficient number of examples. Therefore, the heatmaps presented here are robust enough to indicate the differences (previous studies can also back this claim \cite{schomaker2004automatic2}). To demonstrate the robustness: with the current number of \textit{aleph}s, any pixel of the heatmap with mid-intensity (here, orange with $0.5$ intensity, band $0.46$ to $0.54$, and total intensity being $0-1$) has a probability of $0.05$ for that one pixel to give different results. So even if we were to increase the number of instances of a particular character, the resulting heatmap will not change significantly (it is possible to request heatmaps from all the individual characters by emailing the corresponding author). 

\medskip
\textbf{Step 3}.  After having found statistically significant differences in the neighbourhood structure for columns in the scroll, and after having confirmed that a transition occurs at about the middle of the column series, a more detailed analysis is warranted. Please note, that the actual evidence for the differences comes from the primary and secondary analyses, whereas the current focus is illustrative only. The statistical differences obtained are the result of many small textural and allographic differences. For these allographic differences it is also important to keep in mind that an exhaustive list of possible allographs is not required: an allographic codebook approach will work very well, if it is sufficiently diverse~\cite{schomaker2004automatic}. In the current problem, some of the allographs appear to be more different in their occurrence over the left and right columns, and we can take a look at them for illustrative purposes, while remembering that this concerns partial evidence from the extremes of the distribution. 

\begin{figure}[!ht]
    \centering
    \includegraphics[width=.8\textwidth]{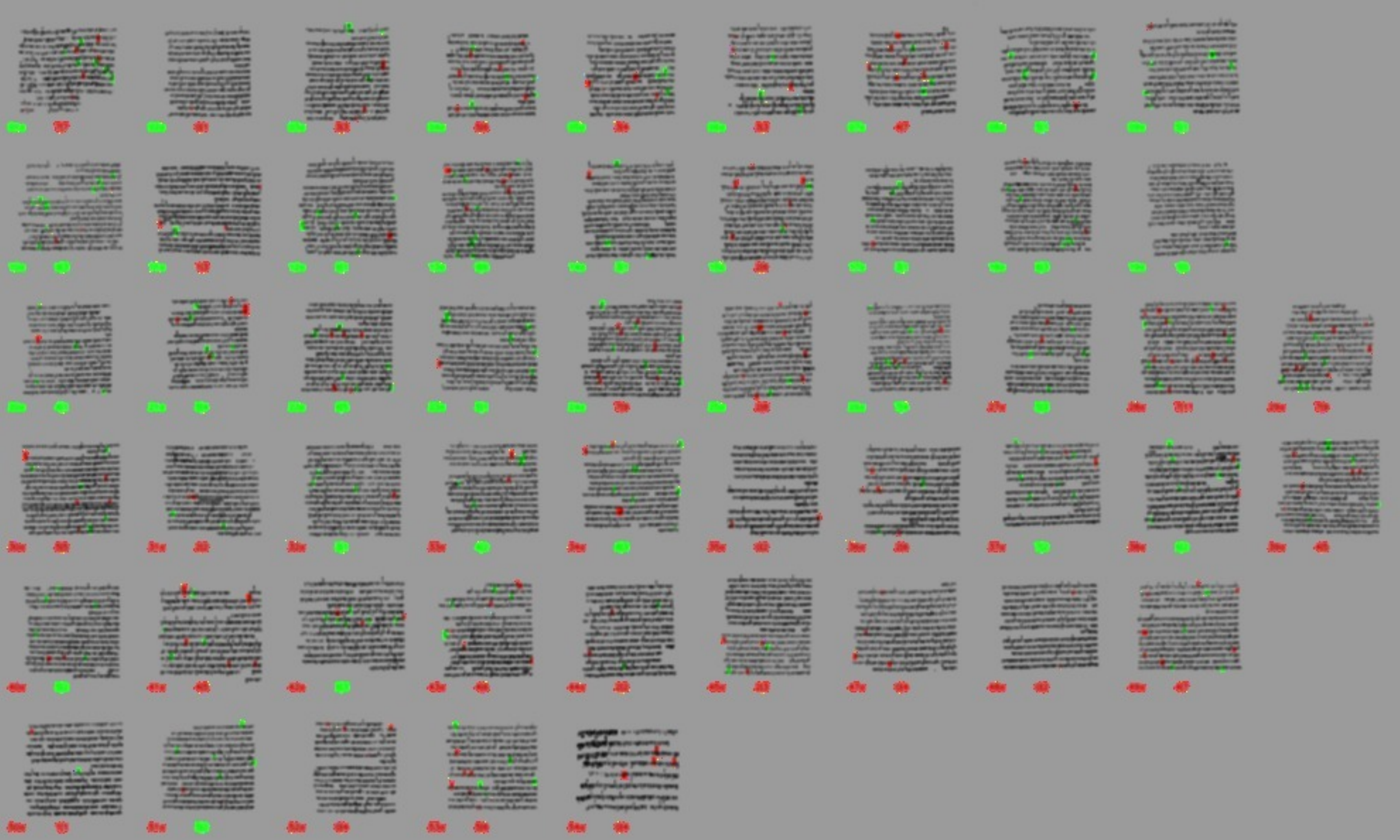}
    \caption{Visually enhanced presence of typical `left' fraglets (green) and `right' fraglets, separately for the `a' split scans (top-halves) of the columns.}
    \label{fig:enhancedfragletsa}
\end{figure}

For the fraglet feature, a selection was made of the most informative fraglets that are able to discriminate between the leftmost ($i<=27$) and the rightmost columns ($i>27$) in the series. Please note, that the number of fraglets in the SOFM is $6400$. From these $6400$ fraglets, we automatically generated sets of fraglets to visualize the differences between the two halves of the manuscript. Thus, we ran tests with thousands of combinations of fraglet sets, each providing a new overview. Figure~\ref{fig:enhancedfragletsa} and Figure~\ref{fig:enhancedfragletsb} show such an overview for the relevant columns (more images can be found in supplementary material \ref{appensec:charts}). Below each column thumbnail, the left blob indicates the ground truth (`left series' is green, `right series' is red), whereas the colour immediately to the right of it shows the colour that the subset of most-informative fraglets predicts. These figures illustrate the statistical view of a separation between the two halves of the manuscript.

\begin{figure}[!ht]
    \centering
    \includegraphics[width=.8\textwidth]{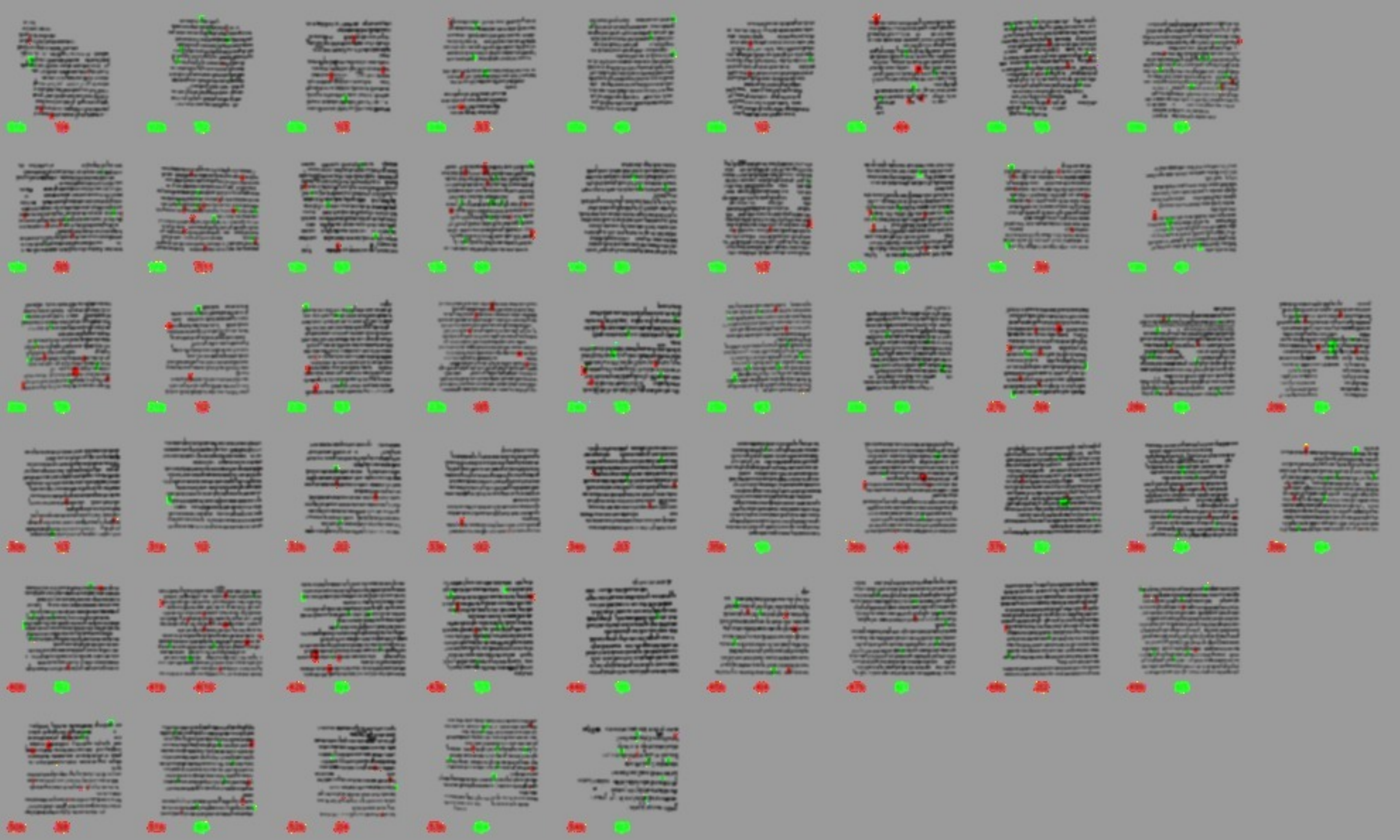}
    \caption{Visually enhanced presence of typical `left' fraglets (green) and `right' fraglets, separately for the `b' split scans (bottom-halves) of the columns.}
    \label{fig:enhancedfragletsb}
\end{figure}

\section{Discussion and conclusions}
The aim of this study was to tackle the palaeographic identification of the unknown scribes of the Dead Sea Scrolls, exemplified by 1QIsa\textsuperscript{a}. The question for 1QIsa\textsuperscript{a} was whether subtle differences in writing should be regarded as normal variations in the handwriting of one scribe or as similar scripts of two different scribes and, if the latter, whether the writing of the two scribes coincides with the two halves of the manuscript. The evidence collection was presented in a chronological manner.

Firstly, an independent observation was made that in feature spaces, the left and right part of the column series, ended up in different regions. Several feature methods confirmed this observation. The preferred explanation is that there were two main scribes responsible for copying 1QIsa\textsuperscript{a}, their work indeed separated between columns 27 and 28 by a three-line lacuna at the bottom of column 27. We see that there is a clear separation between the data points in both the Hinge and the Adjoined feature plot (Fig.\ref{fig:res:hinge-full-color} and Fig.\ref{fig:res:Hinge-fraglet-full}). If we consider an explanation in terms of a large variability within one single scribe, then the question remains why the points are not randomly scattered (between the two sets of columns) on the PCA space in the adjoined feature plot. Instead, there is a clear indication of separation, at least from one of the angles of the plot space. Therefore, a more likely scenario is two different scribes working closely together and trying to keep the same style of writing yet revealing themselves, their individuality, in the textural feature space. 

Secondly, a series of tests was performed on a separate shape feature, a Kohonen map of fragmented contours. A series of five questions was asked, starting with a statistical test whether the pattern of neighbours on the left or right of any given column deviates from the expected random pattern for the case of a single writing style. Because these tests clearly show that the neighbourhood structure is not random, additional analyses were warranted. The distances between columns, as measured in the fraglet-usage space, also showed a highly significant pattern. Finally, the serial column number for the nearest neighbour of each column shows a distinct transition at about the middle of the column series in the scroll. Fitting a logistic model delivered an estimate of the region where this transition occurs, i.e., around column number 27. This point is found without coercion, and emerges from two very different quantitative approaches (a least-squares and a separate Monte-Carlo analysis) on the time series of the column numbers of nearest neighbour matches, for each column. In simple terms: columns on the left clearly tend to yield nearest neighbours on the left, columns on the right clearly tend to yield nearest neighbours on the right. Therefore, these secondary analyses confirm the suspicion raised on the basis of the exploratory primary analyses by other researchers in the team.

Thirdly, our fully-automatic generation of charts with full character shapes for individual Hebrew letters extracted from the digital images of the ancient manuscript of 1QIsa\textsuperscript{a} greatly advances how palaeographic charts have been previously produced, while the subtle differences visible upon close inspection post-hoc of the heatmaps (both thickness and angular differences exist) also show that the use of heatmaps can help to bridge the quantitative analyses and traditional palaeography. Moreover, a post-hoc visual analysis on the most discriminative fraglets in the Kohonen `bag of visual words', which is now allowable given the obtained statistical significance of differences between `left' and `right' in other measures, illustrates the transition point and the differential evidence by colour-marked fraglets in the column images. To be sure, the reverse is also true: if there were no statistical significance of differences between `left' and `right', then it would not have been allowable to look for evidence of difference in post-hoc visual analyses.

Yet, there are at least three variables that we need to be transparent about because these may affect the results in unknown ways. These three variables are: material degradation, writing implements, ink deposition and writing conditions, and limitations on character extraction.

Regarding material degradation, we have to keep in mind that the scrolls, and by extension the images that constitute the data for our pattern recognition and artificial intelligence techniques, have degraded over the centuries and are not anymore in the shape they were once produced. This degradation causes an amount of uncertainty over the derived results, even though we tried our best to extract the original characters using state-of-the-art methods. 

Writing implements and writing conditions can have significant impact on the outcome of the copied scrolls. The use of writing implements could differ in the cutting of the pen's nib and writing conditions could change in the course of time \cite{yardeni2007note}. Although there is no evidence that different writing implements were used in 1QIsa\textsuperscript{a} or a change in writing conditions occurred, the point is that the specific writing implement or a change in writing conditions have an effect on the ink deposition, which in turn affects our modern extraction process of the original characters. 

Finally, regarding limitations on extraction, note that character extraction can never be perfect. Nevertheless, we are confident with our methodology, and it clearly shows excellent extraction results, both qualitatively and quantitatively. Additionally, our feature extraction methods are tested on an independent dataset: `Firemaker image collection for bench-marking forensic writer identification' \cite{Firemaker}. Furthermore, the statistical tests are methodologically robust and independent of the data they are tested on.

The discussion of these variables is not to cast doubt on our study's outcome, which remains inherently sturdy. However, they remind us that the techniques from pattern recognition and artificial intelligence do not give certainty of identification but statistically proven probabilities that can help the human expert understand and decide between different possibilities. One key point to highlight here: this research is by far the most comprehensive and elaborate study on writer identification on historical manuscripts using state-of-the-art computer-based techniques. The use of feature extractions on both macro- and microlevels of character shapes is extensive, gauging a writer's mimetic (cultural) and genetic (bio-mechanical) traits, respectively. The methods used here are rooted in earlier work in forensic writer identification~\cite{schomaker2007using,GIWIS,franke2003wanda,schomaker2008writer}. The minimal use of human interference, the cross-checks and re-validation through statistical tests make this study unique and lay the foundation for future advanced studies.

Thus, the conclusion is that the use of robust pattern recognition and artificial intelligence techniques is a breakthrough for the palaeography of writer identification in the Dead Sea Scrolls. We have demonstrated that despite the near uniform handwriting there is a clear separation between two writing styles in 1QIsa\textsuperscript{a} and that the separation between the two styles coincides with the codicological separation between columns 27 and 28 and the two halves of the manuscript, columns 1–27 and columns 28–54. 

The similarity in handwriting between different scribes can indicate a common training shared by the scribes, perhaps in a school setting or otherwise close social setting, such as in a family context a father having taught a son to write. For five documentary texts it has been suggested that the similarity in script may be the result from a common school training \cite{cotton1997discoveries}. We have otherwise no concrete evidence for such schools but their presence must be presumed \cite{wise2015language,macdonald2017much,healy2018} . 

Our conclusion for 1QIsa\textsuperscript{a} that there were two main scribes also sheds new light on the production of biblical manuscripts in ancient Judea. We have provided new, tangible evidence that such texts were not copied by a single scribe only but that multiple scribes could closely collaborate on one particular manuscript of a text that would come to be regarded and revered as biblical.

\section*{Acknowledgements}
The research for this article was carried out under the ERC Starting Grant of the European Research Council (EU Horizon 2020): The Hands that Wrote the Bible: Digital Palaeography and Scribal Culture of the Dead Sea Scrolls (HandsandBible 640497), principal investigator: Mladen Popović. The authors owe a special debt of gratitude to Eibert Tigchelaar, Drew Longacre, and Gemma Hayes who responded to an earlier draft of this article. For the images of 1QIsa\textsuperscript{a} from the Brill collection we are grateful to Brill Publishers. For the high-resolution, multi-spectral images of the Dead Sea Scrolls we are grateful to the Israel Antiquities Authority (IAA), courtesy of the Leon Levy Dead Sea Scrolls Digital Library; photographer: Shai Halevi. We are very grateful to the staff of the IAA Dead Sea Scrolls Unit for their help and support.

\section*{Author contributions}
\noindent Conceptualization: Mladen Popovi\'{c}, Maruf Dhali, Lambert Schomaker\\
\noindent Data curation: Mladen Popovi\'{c}, Maruf Dhali, Lambert Schomaker\\
\noindent Formal analysis: Mladen Popovi\'{c}, Maruf Dhali, Lambert Schomaker\\
\noindent Funding acquisition: Mladen Popovi\'{c}\\
\noindent Investigation: Mladen Popovi\'{c}, Maruf Dhali, Lambert Schomaker\\
\noindent Methodology: Mladen Popovi\'{c}, Maruf Dhali, Lambert Schomaker\\
\noindent Project administration: Mladen Popovi\'{c}\\
\noindent Resources: Mladen Popovi\'{c}, Lambert Schomaker\\
\noindent Software: Maruf Dhali, Lambert Schomaker\\
\noindent Supervision: Mladen Popovi\'{c}\\
\noindent Validation: Mladen Popovi\'{c}, Maruf Dhali, Lambert Schomaker\\
\noindent Visualization: Mladen Popovi\'{c}, Maruf Dhali, Lambert Schomaker\\
\noindent Writing – original draft preparation: Mladen Popovi\'{c}, Maruf Dhali, Lambert Schomaker\\
\noindent Writing – review and editing: Mladen Popovi\'{c}, Maruf Dhali, Lambert Schomaker\\

% \bibliography{bib}

\begin{thebibliography}{10}

\bibitem{popovic2019}
Popović M.
\newblock The manuscript collections: An overview.
\newblock In Brooke GJ, Hempel C, editors T\&T Clark Companion to the Dead Sea
  ScrollsLondon: T\&T Clark. 2019; p. 37--50.

\bibitem{zahn2017beyond}
Zahn MM.
\newblock Beyond `{Q}umran {S}cribal {P}ractice':{T}he case of the {T}emple
  {S}croll.
\newblock Revue de Qumran. 2017;29/110:185--203.

\bibitem{cotton1997discoveries}
Cotton HM, Yardeni A. Discoveries in the {J}udean {D}esert. Volume 27.
  {A}ramaic, {H}ebrew and {G}reek documentary texts from {N}ahal {H}ever and
  other sites: with an appendix containing alleged {Q}umran texts (The {S}eiyal
  {C}ollection {I}); 1997.

\bibitem{wise2015language}
Wise MO.
\newblock Language and Literacy in Roman Judaea: A Study of the Bar Kokhba
  Documents.
\newblock Yale University Press; 2015.

\bibitem{tov2004scribal}
Tov E.
\newblock Scribal practices and approaches reflected in the texts found in the
  {J}udean {D}esert.
\newblock Brill; 2004.

\bibitem{tigchelaar2003search}
Tigchelaar E.
\newblock In search of the scribe of 1{Q}{S}.
\newblock In: Paul SM, Kraft RA, Schiffman LH, Fields WW, editors. Emanuel:
  Studies in Hebrew Bible, Septuagint, and Dead Sea Scrolls in honor of Emanuel
  Tov. Brill; 2003. p. 339--352.

\bibitem{yardeni2007note}
Yardeni A.
\newblock A note on a Qumran scribe.
\newblock New Seals and Inscriptions, Hebrew, Idumean, and Cuneiform, ed Meir
  Lubetski. 2007; p. 281--292.

\bibitem{ulrich2007identification}
Ulrich E.
\newblock Identification of a Scribe Active at Qumran:
  1{QP}s$^b$-4{QI}sa$^c$-11{QM}.
\newblock Meghillot: Studies in the Dead Sea Scrolls. 2007;6:201--210.

\bibitem{allegro1957dead}
Allegro JM.
\newblock The {D}ead {S}ea {S}crolls. 1957;.

\bibitem{wise1994accidents}
Wise MO.
\newblock Accidents and {A}ccidence: {A} {S}cribal {V}iew of {L}inguistic
  {D}ating of the {A}ramaic {S}crolls from {Q}umran.
\newblock In: Wise MO. Thunder in Gemini and other essays on the history,
  language and literature of Second Temple Palestine. Sheffield: JSOT Press;
  1994. p. 103--151.

\bibitem{alexander2003literacy}
Alexander PS.
\newblock Literacy among {J}ews in {S}econd {T}emple {P}alestine: {R}eflections
  on the {E}vidence from {Q}umran.
\newblock Hamlet on a Hill: Semitic and Greek Studies Presented to Professor T
  Muraoka on the Occasion of his Sixty-Fifth Birthday. 2003; p. 3--24.

\bibitem{golb1990khirbet}
Golb N.
\newblock Khirbet {Q}umran and the {M}anuscripts of the {J}udaean {W}ilderness:
  {O}bservations on the {L}ogic of their {I}nvestigation.
\newblock Journal of Near Eastern Studies. 1990;49(2):103--114.

\bibitem{golb1995wrote}
Golb N.
\newblock Who wrote the Dead Sea scrolls?: The search for the secret of Qumran.
\newblock Scribner; 1995.

\bibitem{crawford2019scribes}
Crawford SW.
\newblock Scribes and scrolls at Qumran.
\newblock William B. Eerdmans Publishing Company; 2019.

\bibitem{sirat2007writing}
Sirat C.
\newblock Writing as handwork: a history of handwriting in {M}editerranean and
  {W}estern culture. 2007;.

\bibitem{papaodysseus2014identifying}
Papaodysseus C, Rousopoulos P, Giannopoulos F, Zannos S, Arabadjis D,
  Panagopoulos M, et~al.
\newblock Identifying the writer of ancient inscriptions and {B}yzantine
  codices. {A} novel approach.
\newblock Computer Vision and Image Understanding. 2014;121:57--73.

\bibitem{davis2007practice}
Davis T.
\newblock The practice of handwriting identification.
\newblock Library. 2007;8(3):251--276.

\bibitem{harralson2017huber}
Harralson HH, Miller LS.
\newblock Huber and Headrick's Handwriting Identification: Facts and
  Fundamentals.
\newblock Crc Press; 2017.

\bibitem{derolez2003palaeography}
Derolez A.
\newblock The palaeography of Gothic manuscript books: From the twelfth to the
  early sixteenth century. vol.~9.
\newblock Cambridge University Press; 2003.

\bibitem{stokes2015digital}
Stokes PA.
\newblock Digital approaches to paleography and book history: some challenges,
  present and future.
\newblock Frontiers in Digital Humanities. 2015;2:5.

\bibitem{kahle1950present}
Kahle P.
\newblock Theologische Literaturzeitung. 1950;75:539.

\bibitem{kahle1951hebraischen}
Kahle P.
\newblock Die hebr{\"a}ischen {H}andschriften aus der {H}{\"o}hle. 1951;.

\bibitem{noth1951note}
Noth M.
\newblock Eine {B}emerkung zur {J}esajarolle vom {T}oten {M}eer.
\newblock Vetus Testamentum. 1951;1(1):224--226.

\bibitem{kuhl1952clerkproperty}
Kuhl C.
\newblock Schreibereigentümlichkeiten: {B}emerkungen zur {J}esajarolle
  ({DSI}a).
\newblock Vetus Testamentum. 1952;2(1):307--333.

\bibitem{brookebisection}
Brooke GJ.
\newblock The {B}isection of {I}saiah in The {S}crolls from {Q}umran.
\newblock In Alexander PS, Brooke GJ, Christmann A, Healey JF, Sadgrove PC,
  editors Studia Semitica: The Journal of Semitic Studies jubilee volume. 2005;
  p. 73--94.

\bibitem{trever1949paleographic}
Trever JC.
\newblock A {P}aleographic {S}tudy of the {J}erusalem {S}crolls.
\newblock Bulletin of the American Schools of Oriental Research.
  1949;113(1):6--23.

\bibitem{ulrich2011discoveries}
Ulrich E, Flint PW.
\newblock Discoveries in the Judaean Desert XXXII: Qumran Cave 1: II. The
  Isaiah Scrolls: Part 2: Introductions, Commentary, and Textual Variants.
\newblock OUP Oxford; 2011.

\bibitem{justnes2015hand}
Justnes {\AA}.
\newblock The Hand of the Corrector in 1QIsa a XXXIII 7 (Isa 40, 7-8): Some
  Observations.
\newblock Semitica. 2015;57:205--210.

\bibitem{brownlee1952manuscripts}
Brownlee WH.
\newblock The manuscripts of {I}saiah from which {DSI}a was copied.
\newblock Bulletin of the American Schools of Oriental Research.
  1952;127(1):16--21.

\bibitem{martin1958scribal}
Martin M.
\newblock The scribal character of the {D}ead {S}ea {S}crolls.
\newblock The scribal character of the Dead Sea scrolls. 1958;.

\bibitem{kutscher1974language}
Kutscher EY.
\newblock The Language and Linguistic Background of the Isaiah Scroll:
  1{QI}sa$^a$. vol.~6.
\newblock Brill; 1974.

\bibitem{giese1988further}
Giese RL.
\newblock Further {E}vidence for the {B}isection of 1{QI}sa$^a$.
\newblock Textus. 1988;14(1):61--70.

\bibitem{cook1989orthographical}
Cook J.
\newblock Orthographical peculiarities in the {D}ead {S}ea biblical scrolls.
\newblock Revue de Qumran. 1989;14(2 (54):293--305.

\bibitem{cook1992dichotomy}
Cook J.
\newblock The {D}ichotomy of 1{QI}sa$^a$.
\newblock Intertestamental essays in honour of J{\'o}sef Tadeusz Milik. 1992;
  p. 7--24.

\bibitem{pulikottil2001transmission}
Pulikottil PU, Pulikottil P.
\newblock Transmission of biblical texts in Qumran: the case of the large
  Isaiah scroll 1{QI}sa$^a$.
\newblock Sheffield Academic Press; 2001.

\bibitem{williamsonscribe}
Williamson H.
\newblock Scribe and {S}croll: {R}evisiting the {G}reat {I}saiah {S}croll from
  {Q}umran.
\newblock In Clines DJA, Richards KH, Wright JL, editors Making a Difference:
  Essays on the Bible and Judaism in Honor of Tamara Cohn Eskenazi. 2012; p.
  329--342.

\bibitem{longacre2013developmental}
Longacre D.
\newblock Developmental {S}tage, {S}cribal {L}apse, or {P}hysical {D}efect?
  1{QI}sa$^a$’s {D}amaged {E}xemplar for {I}saiah {C}hapters 34--66.
\newblock Dead Sea Discoveries. 2013;20(1):17--50.

\bibitem{dhali2017digital}
Dhali MA, He S, Popovi{\'c} M, Tigchelaar E, Schomaker L.
\newblock A digital palaeographic approach towards writer identification in the
  dead sea scrolls.
\newblock In: Proceedings of the 6th International Conference on Pattern
  Recognition Applications and Methods-Volume 1: ICPRAM. vol. 2017. Scitepress;
  Set{\'u}bal; 2017. p. 693--702.

\bibitem{he2017beyond}
He S, Schomaker L.
\newblock Beyond OCR: Multi-faceted understanding of handwritten document
  characteristics.
\newblock Pattern Recognition. 2017;63:321--333.

\bibitem{bulacu2007text}
Bulacu M, Schomaker L.
\newblock Text-independent writer identification and verification using
  textural and allographic features.
\newblock IEEE transactions on pattern analysis and machine intelligence.
  2007;29(4):701--717.

\bibitem{niels2007automatic}
Niels R, Vuurpijl L, Schomaker L.
\newblock Automatic allograph matching in forensic writer identification.
\newblock International Journal of Pattern Recognition and Artificial
  Intelligence. 2007;21(01):61--81.

\bibitem{dhali2020feature}
Dhali MA, Jansen CN, de~Wit JW, Schomaker L.
\newblock Feature-extraction methods for historical manuscript dating based on
  writing style development.
\newblock Pattern Recognition Letters. 2020;131:413--420.

\bibitem{ciula2005digital}
Ciula A.
\newblock Digital palaeography: using the digital representation of medieval
  script to support palaeographic analysis.
\newblock Digital Medievalist. 2005;1.

\bibitem{faigenbaum2016algorithmic}
Faigenbaum-Golovin S, Shaus A, Sober B, Levin D, Na’aman N, Sass B, et~al.
\newblock Algorithmic handwriting analysis of {J}udah’s military
  correspondence sheds light on composition of biblical texts.
\newblock Proceedings of the National Academy of Sciences.
  2016;113(17):4664--4669.

\bibitem{faigenbaum2020algorithmic}
Faigenbaum-Golovin S, Shaus A, Sober B, Turkel E, Piasetzky E, Finkelstein I.
\newblock Algorithmic handwriting analysis of the {S}amaria inscriptions
  illuminates bureaucratic apparatus in biblical {I}srael.
\newblock PLOS ONE. 2020;15(1):e0227452.

\bibitem{shaus2020forensic}
Shaus A, Gerber Y, Faigenbaum-Golovin S, Sober B, Piasetzky E, Finkelstein I.
\newblock Forensic document examination and algorithmic handwriting analysis of
  {J}udahite biblical period inscriptions reveal significant literacy level.
\newblock PLOS ONE. 2020;15(9):e0237962.

\bibitem{KooijG2015}
der Kooij~G V. Classifying early {NW}-{S}emitic scripts: {A} search for writing
  traditions by studying script as artefact. 12 {M}ainz {I}nternational
  {C}olloquium on {A}ncient {H}ebrew (to be published); 29 Oktober--1 November
  2015.

\bibitem{dhali2019binet}
Dhali MA, de~Wit JW, Schomaker L.
\newblock BiNet: {D}egraded-{M}anuscript {B}inarization in {D}iverse {D}ocument
  {T}extures and {L}ayouts using {D}eep {E}ncoder-{D}ecoder {N}etworks.
\newblock arXiv preprint arXiv:191107930. 2019;.

\bibitem{Lim1995brill}
Lim TH, Alexander PS.
\newblock Volume 1.
\newblock In: The Dead Sea Scrolls Electronic Library. Brill; 1995.

\bibitem{dssllnet}
The Leon Levy Dead Sea Scrolls Digital Library;.
\newblock \url{https://www.deadseascrolls.org.il/}.

\bibitem{schomaker2004automatic}
Schomaker L, Bulacu M.
\newblock Automatic writer identification using connected-component contours
  and edge-based features of uppercase western script.
\newblock IEEE Transactions on Pattern Analysis and Machine Intelligence.
  2004;26(6):787--798.

\bibitem{schomaker2007using}
Schomaker L, Franke K, Bulacu M.
\newblock Using codebooks of fragmented connected-component contours in
  forensic and historic writer identification.
\newblock Pattern Recognition Letters. 2007;28(6):719--727.

\bibitem{GIWIS}
Schomaker~LRB BA Bulacu~M. {GIWIS} v3.1 {G}roningen {I}ntelligent {W}riter
  {I}dentification {S}ystem {D}ocumentation V3.1. 10.5281/zenodo.4051006.;
  2011.

\bibitem{bod2003probabilistic}
Bod R, Hay J, Jannedy S.
\newblock Probabilistic linguistics.
\newblock Mit Press; 2003.

\bibitem{kroch1989reflexes}
Kroch AS.
\newblock Reflexes of grammar in patterns of language change.
\newblock Language variation and change. 1989;1(3):199--244.

\bibitem{schomaker2004automatic2}
Schomaker L, Bulacu M, Franke K.
\newblock Automatic writer identification using fragmented connected-component
  contours.
\newblock In: Ninth International Workshop on Frontiers in Handwriting
  Recognition. IEEE; 2004. p. 185--190.

\bibitem{Firemaker}
Bulacu M, Schomaker LRB, Vuurpijl L.
\newblock Writer Identification Using Edge-Based Directional Features.
\newblock In: ICDAR '03: Proceedings of the 7th International Conference on
  Document Analysis and Recognition. Washington, DC, USA: IEEE Computer
  Society; 2003. p. 937--941.

\bibitem{franke2003wanda}
Franke K, Schomaker L, Veenhuis C, Taubenheim C, Guyon I, Vuurpijl L, et~al.
\newblock {WANDA}: {A} generic Framework applied in Forensic Handwriting
  Analysis and Writer Identification.
\newblock HIS. 2003;105:927--938.

\bibitem{schomaker2008writer}
Schomaker L.
\newblock Writer identification and verification.
\newblock In: Advances in Biometrics. Springer; 2008. p. 247--264.

\bibitem{macdonald2017much}
Macdonald M.
\newblock ``Review of {W}ise 2015".
\newblock Journal of Roman Archaeology. 2017;30:832--842.

\bibitem{healy2018}
Healey JF.
\newblock ``Literacy in {L}iterate {S}ocieties’: {T}he {S}cribe in
  {N}abataean and other {A}ramaic {C}ontexts".
\newblock Languages, Scripts, and Their Uses in Ancient North Arabia, ed
  Michael C A Mcdonald. 2018; p. 31--38.

\bibitem{otsu1979threshold}
Otsu N.
\newblock A threshold selection method from gray-level histograms.
\newblock IEEE transactions on systems, man, and cybernetics. 1979;9(1):62--66.

\bibitem{sauvola2000adaptive}
Sauvola J, Pietik{\"a}inen M.
\newblock Adaptive document image binarization.
\newblock Pattern recognition. 2000;33(2):225--236.

\bibitem{gimp}
Team TGD. Gnu image manipulation program - {GIMP} - version 2.8.6; 2016.

\bibitem{dillon1983introduction}
Dillon M. Introduction to modern information retrieval: G. Salton and M.
  McGill.; 1983.

\end{thebibliography}

% Additional material
\newpage
\section*{Supplementary materials}
\newpage
\appendix
\section{Supposed  scribal idiosyncrasies} \label{sec:appen1:scribal}
The study in \cite{ulrich2011discoveries} suggests that there are nine scribal idiosyncrasies in both halves of the manuscript. Since almost all of the features concern scribal practices shared also by other scribes in the scrolls, the nine features listed by \cite{ulrich2011discoveries} are not scribal idiosyncrasies and therefore do not support there being one main scribe in 1QIsa\textsuperscript{a}:
\begin{itemize}
    \item For other examples of writing of parts of words at the end of a line to be repeated in full on the following line for lack of space, see \cite{tov2004scribal}, 107–108;
    \item For other examples of supralinear and infralinear writing at the end of a line, see \cite{tov2004scribal}, 108. Also, 1QIsa\textsuperscript{a} columns 3 and 30 simply did not allow extending much beyond the left column margin because of the stitches connecting two sheets (and in column 45:10 the intercolumn space was already used up);
    \item For a discussion of extending beyond the right column margin, see \cite{tov2004scribal}, 106;
    \item Regarding the ligature \textit{samek} and final \textit{pe} occurring virtually only in \<ksp>: 
        \subitem 1. There are seven examples where this does not occur (cf. 1QIsa\textsuperscript{a} 2:16; 32:17; 33:19; 39:11; 43:17; 45:19; 45:20) and five examples where it does occur (cf. 1QIsa\textsuperscript{a} 7:14; 37:3; 40:15; 45:19; 49:20 [the last one is not listed by \cite{ulrich2011discoveries}]), sometimes in the same line (45:19);
        \subitem 2. The way in which \textit{samek} and final \textit{pe} are written in 1QIsa\textsuperscript{a} 7:14 differs from the other four examples from the second half of the manuscript (especially in the horizontal upper stroke of \textit{samek} and in the horizontal down stroke of final \textit{pe}), whereas those four are written in the same way; 
        \subitem 3. All other occurrences of words with \textit{samek} and final \textit{pe} are non-ligatured except for \<y'sp> in 1QIsa\textsuperscript{a} 49:23;
    \item For another example of starting to write the \textit{lamed} too soon, see 4Q27 15:10;
    \item For many more examples of crossing out words or letters, see \cite{tov2004scribal}, 198–201;
    \item For a discussion of many other examples of cancellation dots, see \cite{tov2004scribal}, 188–198.
\end{itemize}

%%%%%%%%%%%%%%%%%%%%%%%%%%%%%%%%%%%%%%%%%%%%%%%%%%%%%%%%%%%%%%%%%%%%%%%%%%%%%%%%%%%%%%
%%%%%%%%%%%%%%%%%%%%%%%%%%%%%%%%%%%%%%%%%%%%%%%%%%%%%%%%%%%%%%%%%%%%%%%%%%%%%%%%%%%%%%
%%%%%%%%%%%%%%%%%%%% Appendix: B %%%%%%%%%%%%%%%%%%%%%%%%%%%%%%%%%%%%%%%%%%%%%%%%%%%%%
\newpage
\section{Image information} \label{tab:appen1:img}

\begin{table}[h!]
\caption{Brill scan numbers, Shrine of the Book (Israel Museum, Jerusalem) inventory numbers, and their corresponding column numbers for 1QIsa\textsuperscript{a}.}
\begin{tabular}{cc|cc|cc}
\hline \hline
\textbf{Scan number} & \textbf{Column} & \textbf{Scan number} & \textbf{Column} & \textbf{Scan number} & \textbf{Column} \\ \hline \hline
2162/SHR 7001        & \textbf{1}      & 2181/SHR 7022        & \textbf{22}     & 2209/SHR 7043        & \textbf{43}     \\ \hline
2163/SHR 7002        & \textbf{2}      & 2182/SHR 7024        & \textbf{24}     & 2210/SHR 7044        & \textbf{44}     \\ \hline
2164/SHR 7003        & \textbf{3}      & 2183/SHR 7026        & \textbf{26}     & 2211/SHR 7045        & \textbf{45}     \\ \hline
2165//SHR 7004       & \textbf{4}      & 2184/SHR 7027        & \textbf{27}     & 2212/SHR 7047        & \textbf{47}     \\ \hline
2166/SHR 7005        & \textbf{5}      & 2185/SHR 7028        & \textbf{28}     & 2213/SHR 7048        & \textbf{48}     \\ \hline
2167/SHR 7006        & \textbf{6}      & 2186/SHR 7029        & \textbf{29}     & 2214/SHR 7049        & \textbf{49}     \\ \hline
2168/SHR 7007        & \textbf{7}      & 2187/SHR 7030        & \textbf{30}     & 2215/SHR 7050        & \textbf{50}     \\ \hline
2169/SHR 7008        & \textbf{8}      & 2188/SHR 7032        & \textbf{32}     & 2216/SHR 7051        & \textbf{51}     \\ \hline
2170/SHR 7009        & \textbf{9}      & 2189/SHR 7033        & \textbf{33}     & 2217/SHR 7053        & \textbf{53}     \\ \hline
2171/SHR 7012        & \textbf{12}     & 2190/SHR 7034        & \textbf{34}     & 2449/SHR 7010        & \textbf{10}     \\ \hline
2172/SHR 7013        & \textbf{13}     & 2191/SHR 7035        & \textbf{35}     & 2450/SHR 7011        & \textbf{11}     \\ \hline
2173/SHR 7014        & \textbf{14}     & 2192/SHR 7036        & \textbf{36}     & 2451/SHR 7023        & \textbf{23}     \\ \hline
2175/SHR 7015        & \textbf{15}     & 2193/SHR 7037        & \textbf{37}     & 2452/SHR 7025        & \textbf{25}     \\ \hline
2176/SHR 7017        & \textbf{17}     & 2194/SHR 7038        & \textbf{38}     & 2453/SHR 7031        & \textbf{31}     \\ \hline
2177/SHR 7018        & \textbf{18}     & 2195/SHR 7039        & \textbf{39}     & 2455/SHR 7052        & \textbf{52}     \\ \hline
2178/SHR 7019        & \textbf{19}     & 2196/SHR 7040        & \textbf{40}     & 2456/SHR 7054        & \textbf{54}     \\ \hline
2179/SHR 7020        & \textbf{20}     & 2197/SHR 7041        & \textbf{41}     & -                    & -               \\ \hline
2180/SHR 7021        & \textbf{21}     & 2208/SHR 7042        & \textbf{42}     & -                    & -               \\ \hline \hline
\end{tabular}
\end{table}

\newpage
\section{Supplementary material on primary analyses}
\subsection{Preprocessing: Binarization and alignment Correction} \label{appen:preprocess}
Our first step of preprocessing was binarizing the images of 1QIsa\textsuperscript{a}.

It should be noted here that many modern deep-learning methods can be trained end-to-end with the 1QIsa\textsuperscript{a} scroll without performing binarization, but this is not desirable for doing digital palaeography of the scroll. For example, a direct end-to-end solution on clustering the column-images of the 1QIsa\textsuperscript{a} scroll can be achieved for writer identification, but there is always a risk of getting the solution for the wrong cause. For example, the decision of an artificial neural network may be based on spurious correlations with the texture of the parchment. Therefore, it is essential to extract only the ink traces (foreground) and no other material features in the images (background). There are several traditional methods for document binarization. Most commonly used ones are Otsu \cite{otsu1979threshold} and Sauvola \cite{sauvola2000adaptive}. These traditional methods work quite well if the contrast between the ink and the background is relatively large. But for the 1QIsa\textsuperscript{a} images this is not the case mostly due to degradation over time and the skin texture. It is, therefore, important to digitally extract the ink from the parchment. In this study, we use BiNet, a deep-learning-based method especially designed in Groningen to binarize the scrolls images. Instead of using a simple filtering technique, BiNet uses a neural network architecture for the binarization task and therefore yields better output \cite{dhali2019binet}.

After performing the binarization, the images need to be cleaned further. This cleaning is required to get rid of the adjoining columns that appear in each of the images of the target columns. This is an important step as it will ensure that every image corresponding to a particular column should contain the characters from that target column only. Following the cleaning process, rotation and alignment correction needs to be done as well. If the images are rotated with an angle to the horizontal axis, then this affects the feature calculations that are not rotation-invariant. The rotation correction will thus ensure that the lines of texts are aligned horizontally. After this step, in a few cases, minor affine transformation and stretching correction is performed in a restrictive manner. These corrections are also targeted for aligning the texts where the text lines get twisted due to the degradation of the parchment (Fig. \ref{fig:method2}). For the rotation and stretching correction, we have used the well-known GIMP tool, a free and open-source raster graphics editor (version 2.8.16) \cite{gimp}. Finally, we have split each of the columns vertically into half to further validate the tests.

%%%%%%%%%%%%%%%%%%%%%%%%%%%%%%%%%%%%%%%%%%%%%%%%%%%%%%%%%%%%%%%%%%%%%%%%%%%%%%%%%%%%%%
%%%%%%%%%%%%%%%%%%%% Feature extraction %%%%%%%%%%%%%%%%%%%%%%%%%%%%%%%%%%%%%%%%%%%%%%
\subsubsection{Feature extraction: Texture-level} \label{appen:featureText}
Textural methods capture statistical information on attributes of handwriting, like the curvature and slant of the contours of characters. As these methods look at the image as a whole, they do not require a dedicated segmentation technique. The statistical information in the feature vector represents the handwriting style of the document to be used in further analysis. As mentioned above, Hinge is a successful textural feature-extraction technique for the scrolls collection and we use it for 1QIsa\textsuperscript{a} in our current study. Hinge is originally proposed in the work of Marius Bulacu and Lambert Schomaker \cite{bulacu2007text}. 

The Hinge feature is a compact transformation of the handwriting that captures both the writing angle and trace curvature. The bio-mechanics (relative wrist/finger movement control) and allographic choice (the learned and preferred letter shapes) of the writer dictates the slant and roundness of the writing process. As Hinge captures these two basic parameters (slant and roundness), it comfortably succeeds in writer identification and verification.

\begin{figure}[!ht]
    \centering
    \includegraphics[width=.5\textwidth]{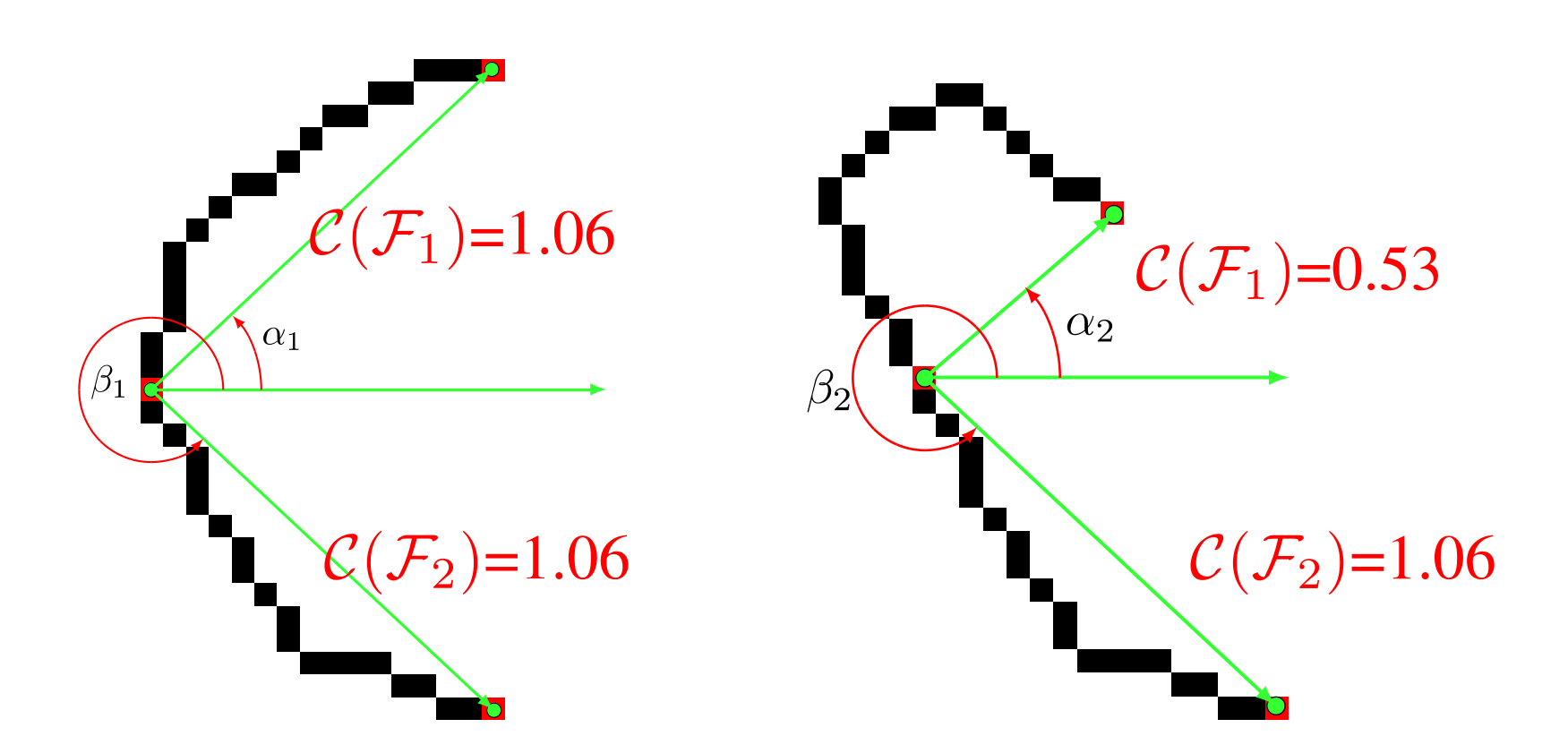}
	\caption{Hinge kernel; the angles and leg-lengths for two different character shapes.}
	\label{fig:app:hinge}
\end{figure}

The Hinge kernel calculates the joint probability distribution of the angle combination of two hinged edge fragments. The joint probability of the orientations is quantized into a two-dimensional histogram $p (\alpha, \beta)$, where the angles $\alpha$ and $\beta (\alpha < \beta)$ are the angles with respect to the horizontal plane, of the two arms of a hinged kernel that is convolved over the edges of a handwritten image. For actual calculations, the hinge can be slid along the contour of each connected ink component of writing. We use $31$ angles ($nbins$: number of bins) for both $\alpha$ and $\beta$ with a length ($nvec$) of $13$ (Fig.\ref{fig:app:hinge}). We only consider the angles that are smaller than $180^{\circ}$ (to get rid of redundancy), and we can exclude the cases in which $\alpha == \beta$ (this is because if they are equal than it implies that they are indicating the same point and there is no useful angle involved). Finally, it results in a feature vector of $465$ dimensions. 

\subsubsection{Feature extraction: Allograph level with neural networks} \label{appen:featureAllo}
The next type of feature we use is on the character-shape (allograph) level, namely the Fraglet. We will briefly explain how the Fraglets are formed. The connected components (mostly the full character shapes) from binarized images are fragmented to get more prototypical shapes from the scrolls collection.

Each fragmented contour (counter-clockwise traced) contains $200$ points yielding $400$ feature values ($x,y$: position of each pixels). The contours are normalized to a centre of gravity at $(0.0,0.0)$, with the radius emanating from that centre being normalized to an average of $r=1.0$. This type of normalization is more stable than bounding-box normalization (Bounding-boxes refer to minimum rectangles containing the ink pixels of individual characters. They create more difficulties in normalization due to different and often arbitrary shapes of ink blobs.). We call these contours the Fraglets.

In order to extract these Fraglets, we used binarized images from the scrolls collection with the condition that the images need to have at least $100,000 (100k)$ black pixels. This ensures automatic selection of images with a relatively high amount of writing. Finally, $746$ full plate images from the scrolls collection fulfilled this criterion. For the full plate images, we use the high-resolution multi-spectral images kindly provided to us by the Israel Antiquities Authority (IAA), which derive from their Leon Levy Dead Sea Scrolls Digital Library project \cite{dssllnet}.

\begin{figure}[!ht]
    \centering
    \includegraphics[width=.8\textwidth]{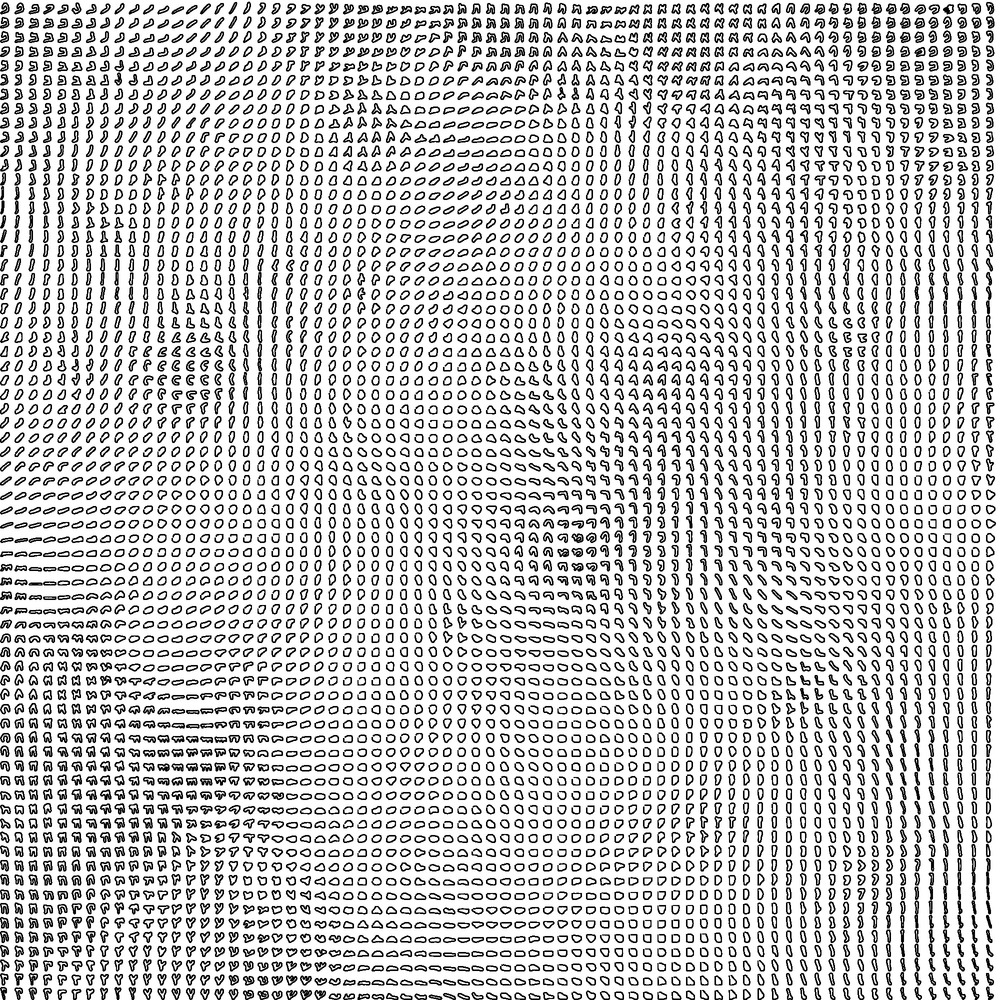}
	\caption{A visualization of 70x70 Kohonen SOFM using Fraglets from the DSS collection.}
	\label{fig:app:allo}
\end{figure}

Using the extracted Fraglets, we than form a Kohonen Map. As mentioned above, this is a self-organizing feature map (SOFM) that uses neural networks with neighbourhood function to preserve the topological properties of the Fraglets. The resulting SOFM contains $70x70$ cells with each cell containing $400$ features (Fig. \ref{fig:app:allo}). The number of cells in the SOFM is derived empirically by measuring the performance on the writer identification data from our previous study on the scrolls \cite{dhali2017digital}. 

Once the Kohonen Map (SOFM) is formed, we then use it to calculate the Fraglet feature for 1QIsa\textsuperscript{a}. For each of the images of the columns, we calculate a feature vector of the output histogram. We take a spread of counts for a Fraglet over $30$ nearest neighbours in the SOFM. We also calculate the cosine feature (and corresponding cosine SOFM file). This involves replacement of the normalized coordinates with cosine/sine pairs. This means $(x,y)$ coordinates become $(cos(\phi),sin(\phi)$) with phi representing the angle with the horizontal axis, for each coordinate point along the contour. Finally, it results in a feature vector of $4900$ dimensions. 

\subsubsection{Adjoined Feature} \label{appen:featureAd}
In order to take advantage from both the textural and allographic feature level, we use a third type of feature namely the Adjoined features. Adjoined features are the weighted combination of both Hinge and Fraglet. The adjoining results in a feature vector of 5365 dimensions preserving the handwriting style description from both feature levels.

\newpage
\section{Supplementary material on secondary analyses} \label{appen3:sec:second}
\subsection{Kohonen map of fragmented connected components}\label{appen3:kohonen:80}
In the so-called bag-of-patterns approach used here, a document is assumed to be characterized by the usage (occurrence) frequencies, i.e., the histogram of the fraglets, similar to the well-known bag-of-words approach in text analysis~\cite{dillon1983introduction}. The distance between such histograms is computed for pairs of document samples. The histogram is assumed to be a feature vector capturing the occurrence of small, prototypical shapes, such that an overall descriptor for the style of each document sample can be computed. Fraglets do not have to correspond to complete characters, they can be smaller or larger than that, and each is mapped to its best-matching centroid in the SOFM, which is guaranteed not to represent an outlier or singleton pattern due to the very large size of the training data.

\begin{figure}[!ht]
    \centering
    \includegraphics[width=.8\textwidth]{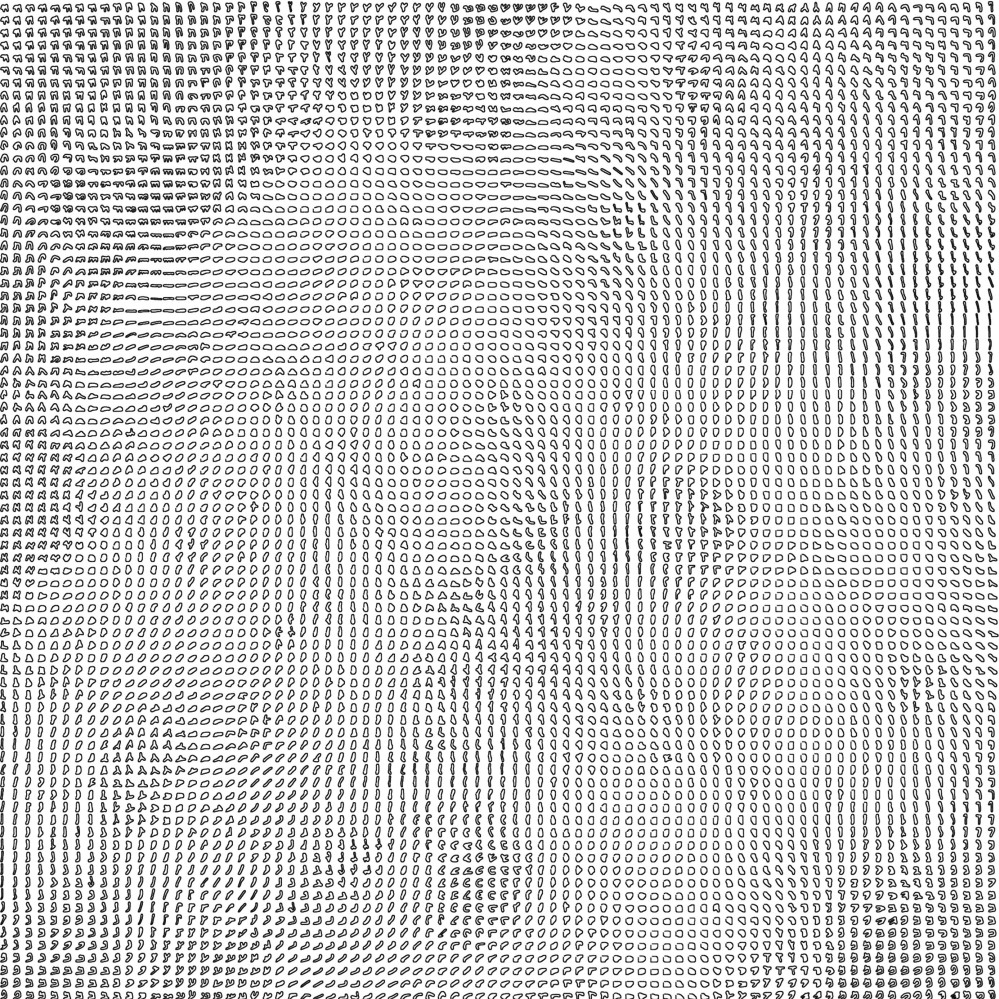}
    \caption{80x80 Kohonen map of fragmented connected components ($200$ x,y points per contour centroid).}
    \label{fig:kohsom80x80}
\end{figure}

\begin{figure}[!ht]
    \centering
    \includegraphics[width=.8\textwidth]{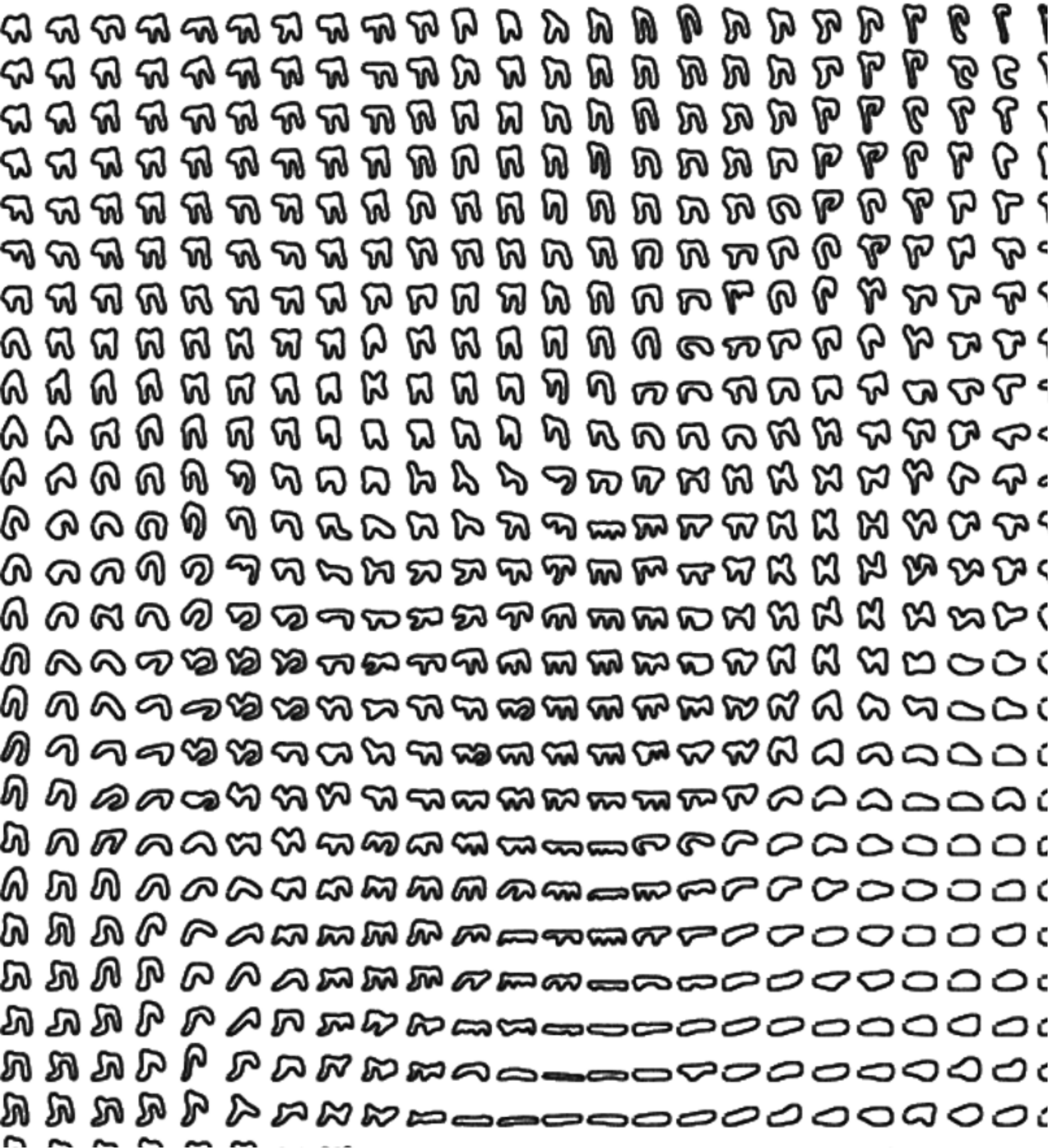}
    \caption{Portion of 80x80 Kohonen map of fragmented connected components.}
    \label{fig:kohsom80x80portion}
\end{figure}

Figure~\ref{fig:kohsom80x80} shows the complete SOFM that was computed, separately from the map that was used in the primary exploratory analysis, on the basis of 600\textit{k} fragmented connected contours derived from binarized IAA images. Figure~\ref{fig:kohsom80x80portion} shows an enlarged portion of the total map.

\subsection{Statistical tests on the fraglet feature distances} \label{appen3:5b}
\medskip
\begin{description}
\item[Step 5a] - If the style were uniform, there should be no difference between the number of times a hit (nearest neighbour) is found on the left or on the right of a column number. Using the Chi-square test, the deviation from the expected frequencies can be computed. If it is more likely that a point on the left has a hit on the left in the sequence, or, vice versa, finding a hit on the right of a point that is on the right, then the distribution is not homogeneous. The window under consideration in the column series is varied from 9 to 26: big enough to catch hits, but smaller than the mid point of the sequence. The returned probability that the pattern of counts is non-accidental will be averaged and a graph will be plotted over the column numbers. A minimum or dip in the curve will be indicative of a column number where the voting pattern for left vs right column hits is not random. The common threshold of $\alpha = 0.05$ will be used to decide for such singular points. No information concerning a critical column number is used. Due to the dependent nature of the running time window of left and right votes for neighbors, additional testing is needed.

\item[Step 5b] - If the style were uniform over columns, the distances to the nearest neighbours on the left and right should be comparable (of similar value) over the column series. On the other hand, if there are style differences, the average value of the distance may change over the column series. For this, a one-way analysis of variance can be used, or a t test, with the categories left and right, for the leftmost and rightmost columns in the series, respectively. Also here, a windowed approach is used, where distances are computed over windows of size 18 to 26 columns and averaged. Please note that, similar to the approach in Question 5a, no information is used concerning a column where style transition may be supposed to occur.

\item[Step 5c] - If the style were uniform, we would expect the same average position for hits over the column series. Indeed, the average position should be in the middle of the column series. On the other hand, if there are style differences, the average estimated position per column would vary. In the case of a linear style development in the series, the estimated average position of hits would also vary linearly. If a  sudden change in style occurs, alternatively, we would expect something like a ‘step response’, i.e., a discontinuity in the series.

\item[Step 5d] - Following 5c, if there is a phase transition in the sequence of columns, fitting a logistic curve on the variable 'average neighbour position' over columns should reveal the switching point reliably, i.e., with a high Pearson correlation of the fit. The number of the critical phase-transition column is the output of this test.
\end{description}

\begin{verbatim}
________________________________________________________
One-way ANOVA
Name             N     Mean       SD      Min      Max 
Left            18    0.238    0.003    0.233    0.243 
Right           17    0.231    0.008    0.220    0.249 
Total           35    0.234    0.007    0.220    0.249 
________________________________________________________
Weighted Means Analysis:
Source           SS    df         MS        F       p
Between       0.000     1      0.000   11.189   0.002 **
________________________________________________________
t-test
Name          N     Mean       SD      Min      Max 
Group-1      18    0.238    0.003    0.233    0.243 
Group-2      17    0.231    0.008    0.220    0.249 
Total        35    0.234    0.007    0.220    0.249 
________________________________________________________
Weighted Means Analysis:
t(33) = 3.345   p = 0.002
\end{verbatim}

The distance of a column with a nearest neighbour is significantly different between matches found to the left and the right $(p < 0.005)$, with a slightly larger distance for the first half $(d=0.238)$ as compared to the second half $(d=0.231)$. One-way anova and a t-test both return $p=0.002$.

The average distances obtained if the queried column is on the left of the found column in the sequence (best hit is in the future). Between column 25 and 30, this distance drops, i.e., ‘future’ columns fit better. After column 35, the distance increases again.The mirror version, i.e., ‘best hit for a query is in the past’ also shows a transition between column 25 and 30. The pattern is a bit less clear but confirms the notion of an accident. Average column position of the best fitting neighbour for a column. Average position of best-fitting neighbours of a column, in the column series. Left of column 27, the average position  of the hits is between 20-25. On the right of column 27, the average position of hits is between 30-35. The light blue line represents column 27. In case of a linear style development, the diagonal blue line should have been approximated. In case of no style development, the y-values should have been about constant.

% 5d significance test on avgpos
% Schomaker, L.R.B. <l.r.b.schomaker@rug.nl>
% Fri, Aug 28, 9:08 PM
% to Popovic, M.A.

One-way ANOVA for variable: position of nearest neighbour (column number) of a given DSS column. The nearest neighbour is computed in fraglet-histogram space ($Ndim=6400$), for two groups: left (column $<=$ 27) and right (column $>$ 27).

\begin{verbatim}
________________________________________________________
Name          N     Mean       SD      Min      Max
LEFT          27   23.847    4.008   16.312   35.688
RIGHT         27   32.023    3.689   25.625   38.688
Total         54   27.935    5.620   16.312   38.688
________________________________________________________
Weighted Means Analysis:
Source           SS    df         MS        F     p
Between     902.418     1    902.418   60.822 0.000 ***
Within      771.519    52     14.837
________________________________________________________
Again: p < 0.001
\end{verbatim}

Also this analysis indicates that the between-column similarity is highest 'ipsilateral' with respect to the cut point (column 27):  left looks like left, right looks like right. The significance is so high that with three decimals, p appears as zero in the output of the statistics tool. We can safely say that $p < alpha = 0.001$. This is a rigorous alpha, similar to medical sample comparisons, i.e., a three stars result (***). In other words: the probability that this difference is the consequence of random fluctuation, is less than $p=0.001$.
\begin{itemize}
    \item The midpoint for category 'left' is at column 24.
    \item The midpoint for category 'right' is at column 32.
\end{itemize}

\subsubsection{Finding the phase transition using a logistic fit}
Assumption: in the column series there is a phase transition somewhere in the series. Indications for this came from Chi-square tests on the distributions of hits left|right of a target column. These tests indicates a switch at around column 27. When using this as the split criterion, a subsequent one-way anova revealed significant difference p < 0.001 for columns on the left and right, who appear to have their nearest-neighbour hits on the left and right, respectively, consistent with the expectation of large similarity within a grouping. If we model the series as a phase transition, using a logistic function, will that transition occur at or about column 27? 

Without seeding a logistic function estimator with knowledge  concerning the magical number 27, this was the output:

\begin{verbatim}
__________________________________________________________________
                        xoff       yoff       amplitude  steepness
mc-logist-reg-predict   27.824583  24.094666  7.983507   0.924704 
__________________________________________________________________
\end{verbatim}

The value of xoff means that the transition column is estimated to occur between column 27 and 28, with a transition steepness that is smooth: In the separate .svg plot it lasts from 24-32.

Sigmoid regression analysis output:
\begin{verbatim}
______________________________________________________
Analysis for 54 cases of 2 variables:
Variable      sigmoid     avgpos
Min           24.0947    16.3125
Max           32.0782    38.6875
Sum         1514.0753  1508.5000
Mean          28.0384    27.9352
SD             3.8642     5.6199
______________________________________________________
Correlation Matrix:
sigmoid        1.0000
avgpos         0.7432     1.0000
Variable      sigmoid     avgpos
______________________________________________________
Regression Equation for sigmoid:
sigmoid  =  0.511 avgpos  +  13.7632
______________________________________________________
Significance test for prediction of sigmoid
    Mult-R  R-Squared      SEest    F(1,52)   prob (F)
    0.7432     0.5523     2.6102    64.1608     0.0000
______________________________________________________
\end{verbatim}

The fit of the sigmoid transition model is significant, with a correlation r that equals 0.74 (p < 0.001). An exact fit would have yielded r=1.0. 
Although not perfect, r=0.74 would be considered as a very robust correlation in psychology and biology. 

The model would explain 55\% of the variance in the data, which is not strange, given the fact that the model is a stylized description
of a time sequence with irregularities. If we smooth the  irregularities over time, using a running average over a limited 3 or 
over 5 samples (columns), to smooth out the within writer variation, the correlation with the sigmoid increases considerably:
\begin{itemize}
    \item If we smooth the column time series over 3 values, r=0.87  (76\% var. explained variance by sigmoid phase transition)
    \item If we smooth the column time series over 5 values, r=0.93 (86\% var. explained variance by sigmoid phase transition)
\end{itemize}

\subsection{Least-squares fitting (Scipy)  of a logistic curve on the estimated average serial position of the nearest-neighbour of a column}

\begin{figure}[!ht]
    \centering
    \includegraphics[width=.7\textwidth]{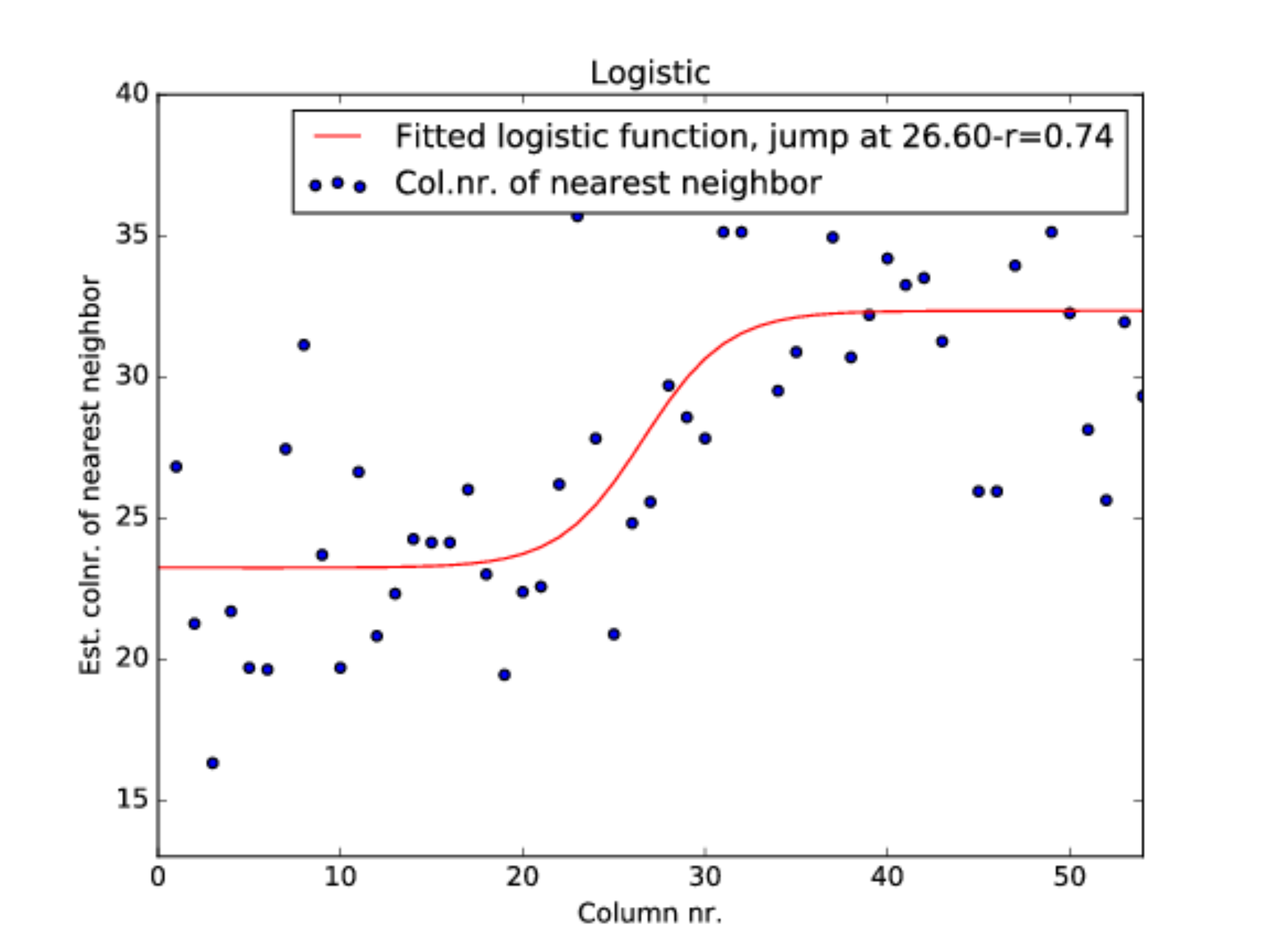}
    \caption{Average serial position of nearest-neighbour of a column in fraglet-feature space (raw values), with a least-squares fitted logistic curve.}
    \label{fig:scipyfitlogisticraw}
\end{figure}

\begin{figure}[!ht]
    \centering
    \includegraphics[width=.7\textwidth]{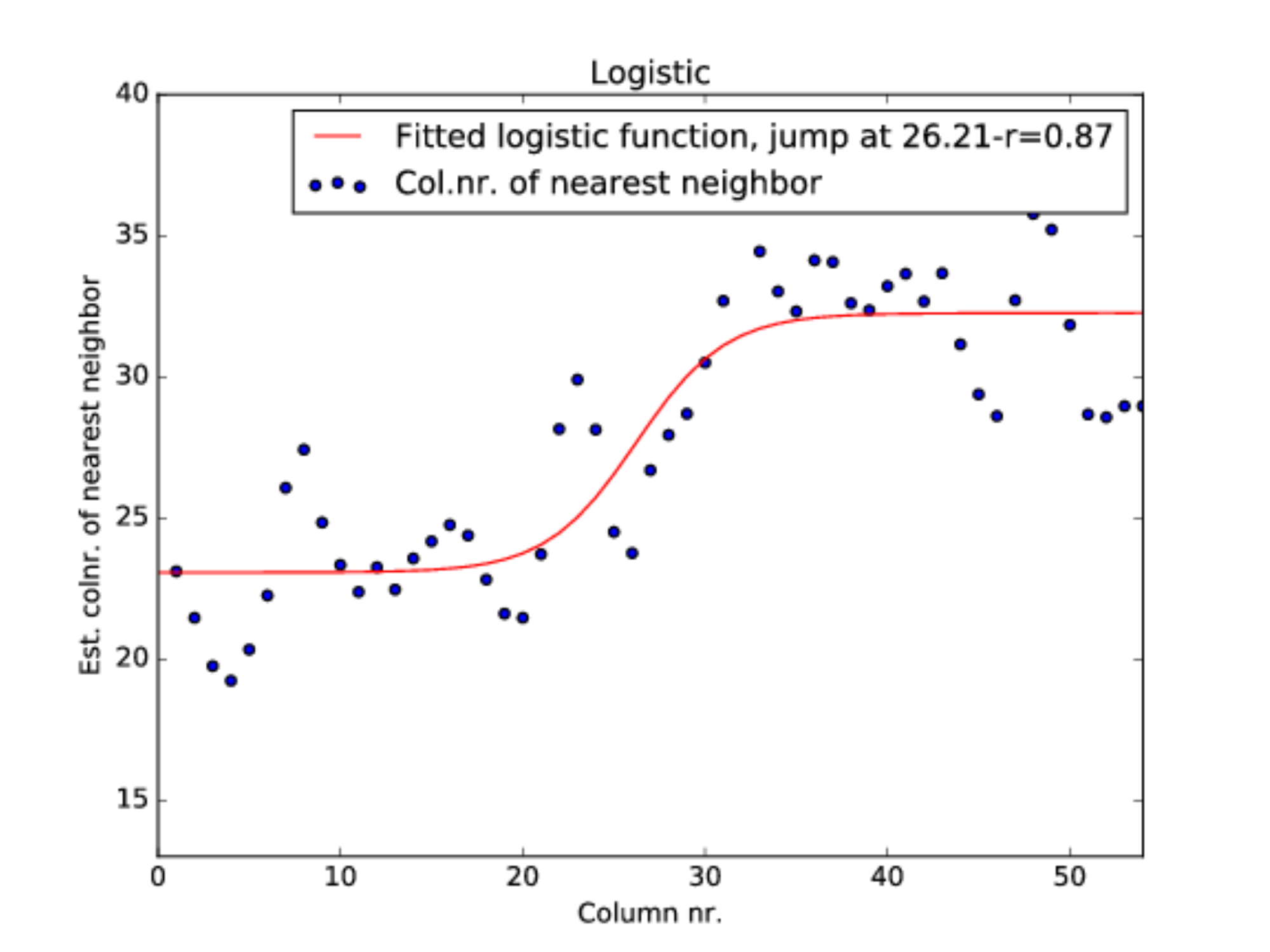}
    \caption{Average serial position of nearest-neighbour of a column in fraglet-feature space (smoothed over 3 values), with a least-squares fitted logistic curve.}
    \label{fig:scipyfitlogisticsmo3}
\end{figure}

\begin{figure}[!ht]
    \centering
    \includegraphics[width=.7\textwidth]{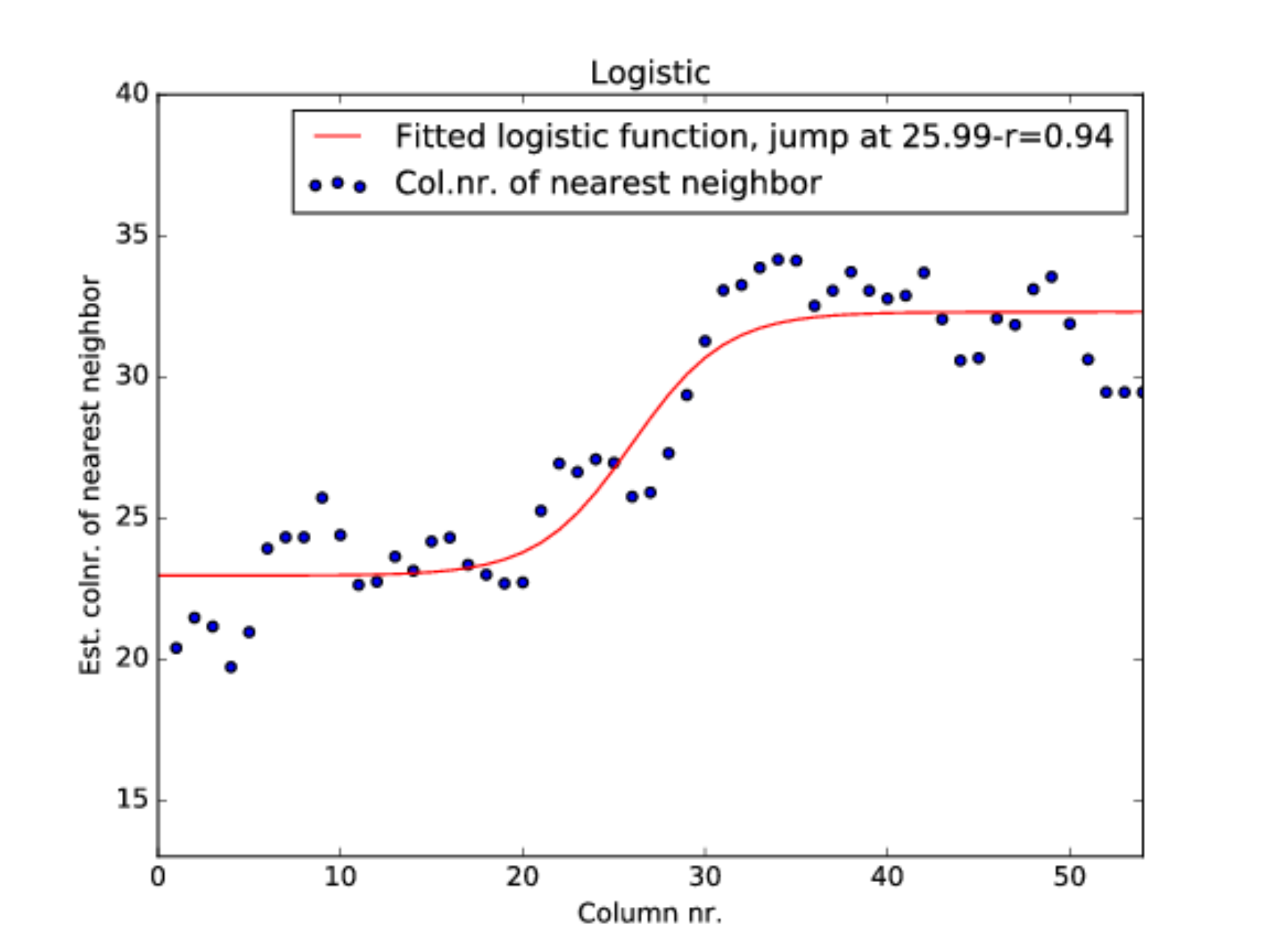}
    \caption{ Average serial position of nearest-neighbour of a column in fraglet-feature space (smoothed over 5 values), with a least-squares fitted logistic curve.}
    \label{fig:scipyfitlogisticssmo5}
\end{figure}

The result of the Monte Carlo-based logistic model fit was replicated with a more traditional least-squares curve fit (Python scipy), yielding a phase transition at column 26.6 for raw data, with r=0.74, at 26.2 for a smoothed time series with a window of three points (r=0.87) and a transition at column 26 for a smoothed time series with a window of five points (r=0.94). Curve-fitting results are shown in Fig.~\ref{fig:scipyfitlogisticraw}, \ref{fig:scipyfitlogisticsmo3} and~\ref{fig:scipyfitlogisticssmo5}.  Even without smoothing, the phase transition is clearly visible. With smoothing, the pattern is  even more clear (window size 5, Fig.~\ref{fig:scipyfitlogisticssmo5}).

\begin{verbatim}
_________________________________________________________________
Raw Monte Carlo (mc:) and least squares (scipy:) results
_________________________________________________________________
  Xoffset     Yoffset   A       steepness
 27.824583 24.094666 7.983507 0.924704 < mc:try     r=0.7432
 28.243372 24.213305 7.994352 2.602412 < mc:RAW     r=0.7363
 27.727867 23.932599 8.002640 0.876055 < mc:RAW     r=0.7432
 27.634910 24.354421 7.740445 1.295146 < mc:RAW     r=0.7414
 27.882597 24.194399 7.872786 7.975232 < mc:RAW     r=0.7365
 26.605149 23.233309 9.102090 0.435106 < scipy:RAW  r=0.7441
_________________________________________________________________
 27.9197 Avg,RAW

 27.170642 23.515902 7.901666 0.751979 < mc:SMO3    r=0.8692
 25.067162 22.548375 9.256635 0.362236 < mc:SMO3    r=0.8713
 24.622219 22.198244 9.320780 0.468216 < mc:SMO3    r=0.8665
 28.309579 23.505822 8.146873 1.383725 < mc:SMO3    r=0.8674
 26.210565 23.064034 9.189271 0.407289 < scipy:SMO3 r=0.8728
_________________________________________________________________
 26.2924 Avg,SMO3

 25.907792 23.278395 8.809249 0.572107 < mc:SMO5    r=0.9316
 23.054408 22.920690 8.547156 0.335815 < mc:SMO5    r=0.9243
 24.057952 22.826461 8.236999 0.341474 < mc:SMO5    r=0.9309
 24.405605 22.058579 9.809585 0.280687 < mc:SMO5    r=0.9319
 25.990002 22.951100 9.342036 0.389834 < scipy:SMO5 r=0.9360
_________________________________________________________________
 24.3564 Avg,SMO5
\end{verbatim}

The least-squares approach gives one result. The Monte Carlo estimation is done a few times (1 hour of computing per fit). Although smoothing over 5 columns (SMO5) gives the highest value of the Pearson correlation, the smoothing also biases the estimation of the transition point. Therefore the estimate of the transition point for 'RAW' data is to be preferred. The estimation yields a negative number for the Xoffset, this corrected here, to be consistent with Eq. 1.
\clearpage
\hfill
\clearpage

\newpage
\section{Tertiary analyses} \label{appensec:charts}
\begin{figure}[!ht] 
    \centering
    \includegraphics[width=\textwidth]{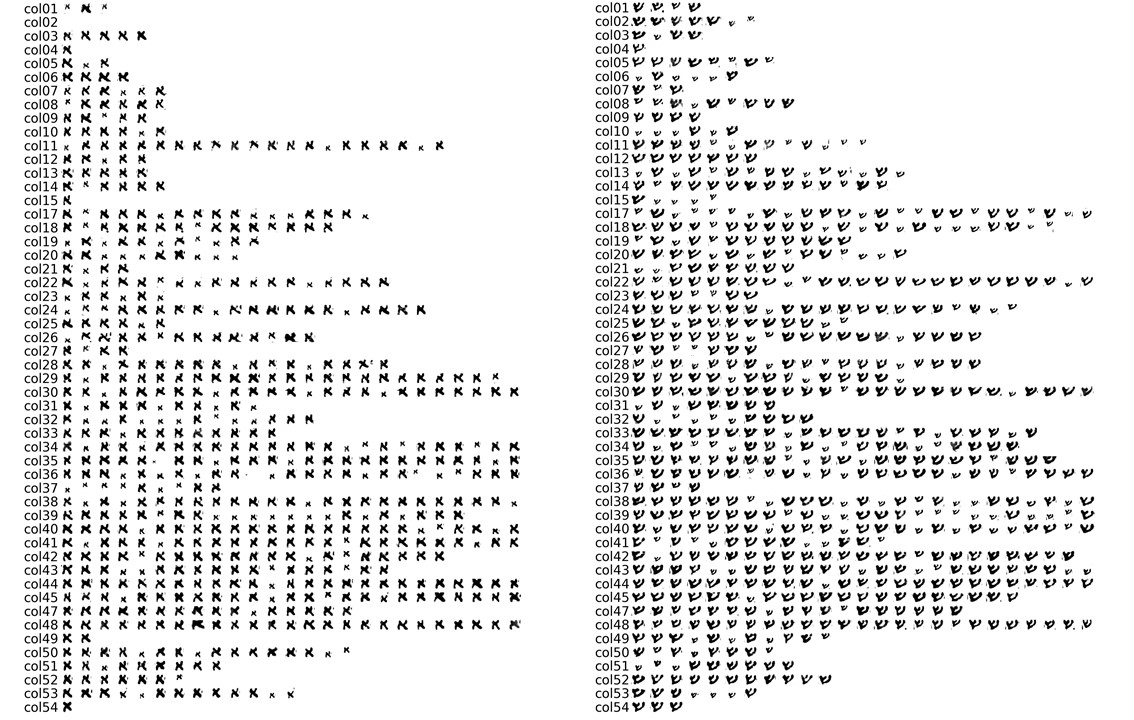}
	\caption{Individual character shapes of \textit{aleph} (left) and \textit{shin} (right) extracted from each of the columns of 1QIsa\textsuperscript{a}.}
	\label{fig:app:charts}
\end{figure}

The charts with full character shapes for individual Hebrew letters improve significantly on the traditional palaeographic chart. Each instance of a character can be directly traced back to its exact position in the manuscript of 1QIsa\textsuperscript{a}. Also, there is no modern human hand involved, either in retracing the characters or in character reconstruction. The ink traces are extracted as is from the digital images and retain the movements once made by the ancient scribe's hand. Figure \ref{fig:app:charts} presents two such charts. It is possible to request charts from all the individual characters by emailing the corresponding author. 
\begin{figure}[!ht]
    \centering
    \includegraphics[width=.8\textwidth]{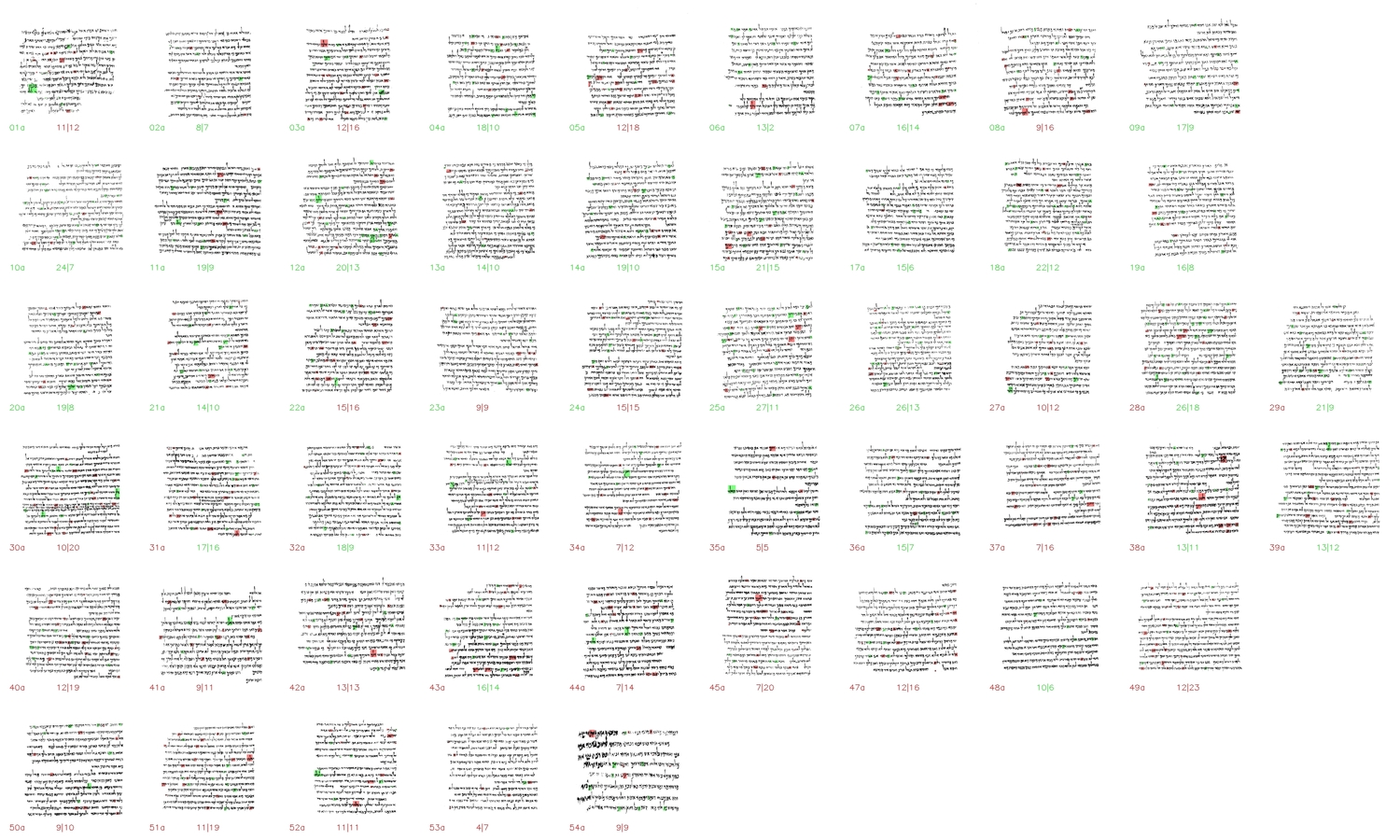}
    \caption{Visually enhanced presence of typical `left' fraglets (green) and `right' fraglets, separately for the `a' split scans (top-halves) of the columns.}
    \label{fig:enhancedfragletsa:appen}
\end{figure}
\clearpage
\begin{figure}[!ht]
    \centering
    \includegraphics[width=.8\textwidth]{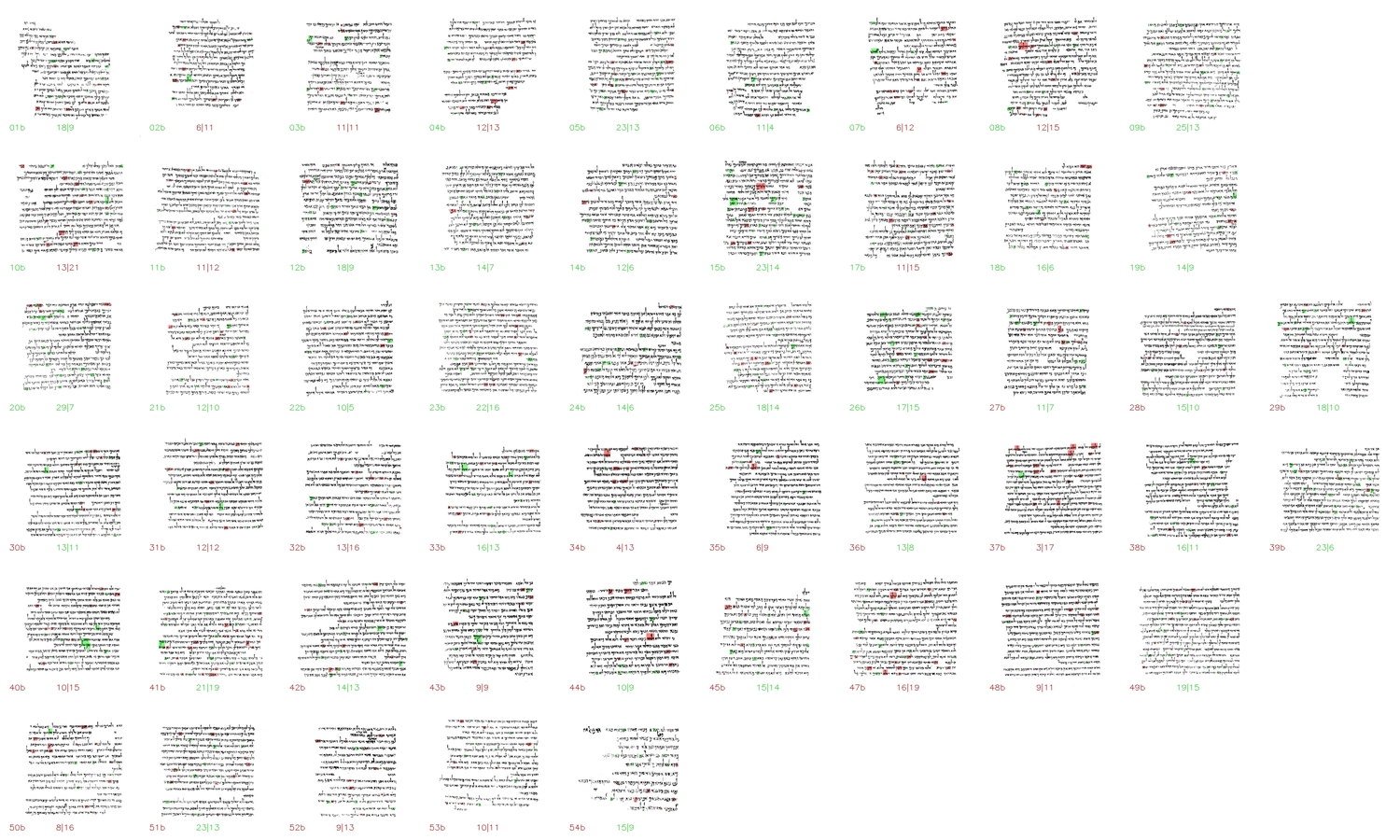}
    \caption{Visually enhanced presence of typical `left' fraglets (green) and `right' fraglets, separately for the `b' split scans (bottom-halves) of the columns.}
    \label{fig:enhancedfragletsb:appen}
\end{figure}
\hfill

\end{document}